\documentclass[lettersize,journal]{IEEEtran}
\usepackage{graphicx}
\usepackage{amssymb}
\usepackage{booktabs} % 导入booktabs宏包
\usepackage[colorlinks,linkcolor=blue,citecolor=blue,urlcolor=blue]{hyperref}
\usepackage{colortbl}
\usepackage{multirow}%表格跨行
\usepackage{amsmath}%数学公式
\usepackage{setspace}%间距宏包
\usepackage{bbm}
\usepackage[utf8]{inputenc}
\usepackage{url}
\usepackage{amssymb}
\usepackage{bbding}
\usepackage{pifont}
\usepackage{wasysym}
\usepackage{utfsym}
\usepackage{fontawesome}
\usepackage{caption}
\usepackage{algorithm}
\usepackage{algorithmic} 
\usepackage{amsmath}  
\usepackage{xcolor}
\usepackage{color}
\newcommand{\red}[1]{\textcolor{red}{#1}}
\newcommand{\blue}[1]{\textcolor{blue}{#1}}
\begin{document}
%\begin{frontmatter}



\title{Triple-domain Feature Learning with Frequency-aware Memory Enhancement for Moving Infrared Small Target Detection} 
%\title{Frequency-aware Memory Enhancement in Dual Domains for Moving Infrared Small Target Detection} 
\author{Weiwei Duan, Luping Ji$^\ast$, \IEEEmembership{Member, IEEE}, Shengjia Chen, Sicheng Zhu, Mao Ye, \IEEEmembership{Member, IEEE}
\thanks{$^\ast$Corresponding author: Luping Ji.}
\thanks{This work is supported by the Aeronautical Science Foundation of China (ASFC) under Grant No. 2022Z071080006 and the National Natural Science Foundation of China (NSFC) under Grant No. 61972072.}
\thanks{Weiwei Duan, Luping Ji, Shengjia Chen, Sicheng Zhu and Mao Ye are with the School of Computer Science and Engineering, University of Electronic Science and Technology of China, Chengdu 611731, China (e-mail: dww@std.uestc.edu.cn;
	jiluping@uestc.edu.cn; 
    shengjiachen@std.uestc.edu.cn;
    sichengzhu@std.uestc.edu.cn;
    maoye@uestc.edu.cn)} }
% \thanks{Xiaoyong Yao is with the School of Mechanical and Electrical Engineering, Jinggangshan University, Ji'an 343009, China (e-mail: yaoxiaoyong@jgsu.edu.cn)} 

%Elsevier
% \author{Weiwei Duan, Luping Ji, Shengjia Chen, Mao Ye, Xiaoyong Yao}
% \ead{dww@std.uestc.edu.cn}
% \affiliation{organization={School of Computer Science and
% Engineering,University of Electronic Science and Technology of China},
%             %addressline={Hezuo Street, High-tech Zone}, 
%             city={Chengdu},
%             postcode={611731}, 
%             %state={SiChuan},
%             country={China}}

\maketitle
\begin{abstract}
As a sub-field of object detection, moving infrared small target detection presents significant challenges due to tiny target sizes and low contrast against backgrounds. 
%问题
Currently-existing methods primarily rely on the features extracted only from spatio-temporal domain. 
Frequency domain has hardly been concerned yet, although it has been widely applied in image processing.
%\red{often failing to capture the intricate characteristics of small moving targets thoroughly.}
%For further enhancing feature representation, more information domains such as frequency are believed to be potentially valuable.  
%, generally neglecting the frequency information in images.
%However, this scheme could suffer from information loss owing to complex dynamic scenes and motion blur caused by the fast motion of targets or cameras.
%Recently, frequency domain transformation has become a focus in computer vision because it provides rich perspectives to help understand image features and structures.
To extend feature source domains and enhance feature representation, we propose a new \emph{Triple-domain Strategy} ({Tridos}) with the frequency-aware memory enhancement on spatio-temporal domain for infrared small target detection. 
%that incorporates frequency-domain analysis to solve these problems.
% 亮点
%Specifically, we develop
In this scheme, it effectively detaches and enhances frequency features by a local-global frequency-aware module with Fourier transform. Inspired by human visual system, our memory enhancement is designed to capture the spatial relations {of infrared targets among} video frames.
Furthermore, it encodes temporal dynamics motion features via differential learning and residual enhancing.
% A temporal dynamics encoding module (TDEM) based on differential learning and residual enhancement is proposed to robustly encode motion features. 
% Additionally, we introduce a local-global frequency-aware block (LGFM) that utilizes the Fourier transform to capture frequency features. 
Additionally, we further design a residual compensation to reconcile possible cross-domain feature mismatches.
%cross domains and have developed a new multi-frame infrared small target dataset (ITSDT-15K), given the lack of MISTD datasets.
% 优势
To our best knowledge, {proposed} Tridos is the first work to explore {infrared} target feature learning comprehensively in spatio-temporal-frequency domains.  
The extensive experiments on three datasets ({i.e.,} DAUB, ITSDT-15K and IRDST) validate that our triple-domain {infrared feature} learning scheme could often be obviously superior to state-of-the-art ones.
Source codes are available at https://github.com/UESTC-nnLab/Tridos.
%解决思路
% In this paper, we propose a frequency-aware memory enhancement network (FMENet) that utilizes spatial-temporal and frequency domain features to solve the problems.
\end{abstract}
% \begin{highlights}
% \end{highlights}
%\end{frontmatter}
\section{Introduction}
\IEEEPARstart{I}{nfrared} small target detection (ISTD) has the advantages of being less negatively affected by the environment and {being} highly resistant to external electromagnetic interference \cite{zhao2022survey}. It holds significant application value in some military {and civil} fields, such as {early invasion} warning, infrared guidance and maritime rescue \cite{kou2023survey}.
{Although ISTD has been developing rapidly in recent years, it still faces huge challenges. First, due to their tiny size, infrared targets could often lack some obvious visual features, such as shapes and textures. Second, with a large detection range, targets are often characterized by low contrast and low signal-to-noise ratio (SNR).
Third, the fast motion of targets and background noise interference usually result in severe infrared 
%images that often exhibit severe 
background fluctuations, and target boundary blurring.
%usually has blurred boundaries.
}
% However,  Furthermore,
% Due to the tiny size of targets, they could often lack obvious visual features and have blurred boundaries, characterized by low contrast and low signal-to-noise ratio (SNR).
Consequently, accurately locating and tracking moving small targets in infrared images and videos has been {an attractive} research area in computer vision, full of research challenging.
\begin{figure}[t]
    \centering
\includegraphics[width=\linewidth]{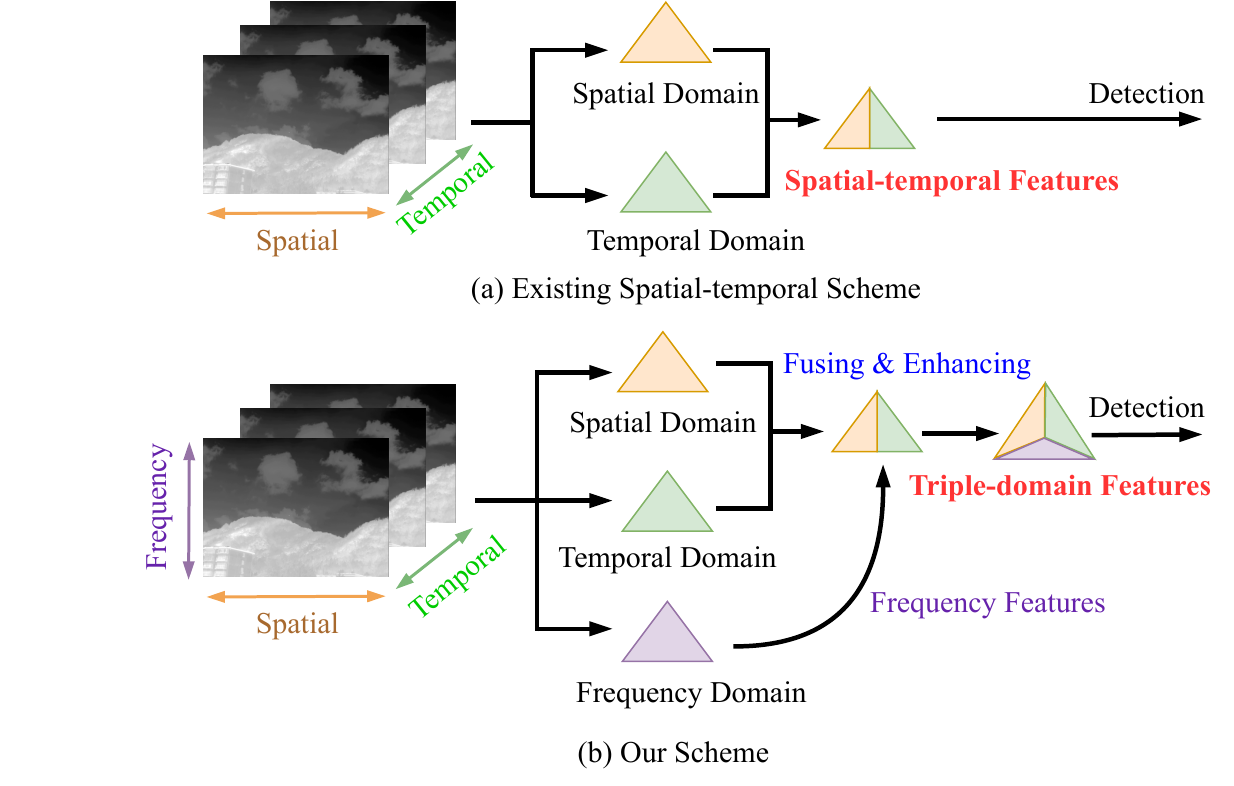}
    \caption{
    The comparison between existing spatio-temporal scheme and our triple-domain learning scheme. Our scheme extracts features in spatio-temporal-frequency domains.
    % The comparison between different schemes of moving infrared small target detection. Our scheme extract features both spatial-temporal and frequency domains.
    }
    \label{fig:intro}
\end{figure}

Over the past decades, many detection approaches have been proposed. 
% to detect infrared small targets. 
From the {difference} of research {strategies}, they could usually be categorized into model-driven and data-driven {methods} \cite{IRTransDet}. 
% Model-driven methods leverage the image characteristics by deeply analyzing the differences between targets and backgrounds \cite{Tophat,chen2013lcm,wang2021NPSTT}. 
Model-driven methods deeply analyze image characteristics to distinguish targets from backgrounds, often {by} contrast and texture differences \cite{Tophat,chen2013lcm}.
Although these methods have achieved {impressive} results, they
heavily depend on prior knowledge and hand-crafted features, lacking learning ability \cite{wang2021NPSTT}. 
% By comparison, data-driven methods can adaptively learn target features on a large number of training samples \cite{NUAAACM,AGPCNet,RDIAN},  which have become the mainstream.
{
Recently, artificial intelligence technologies, especially deep learning, have been developing rapidly, e.g., the SpectralGPT \cite{SpectralGPT}, LRR-Net \cite{LRRNet} and CasFormer \cite{CasFormer}. These methods show impressive promise, due to their efficient capabilities of feature learning.  
% which shows incredible promise in remote sensing applications. 
% For example, \cite{SpectralGPT} proposes the SpectralGPT, a spectral remote sensing foundation model for various visual tasks.
% LRR-Net \cite{LRRNet} introduces a novel framework for integrating low-rank representations and deep learning techniques.
% Moreover, \cite{CasFormer} employs cascaded transformers for fusion-aware computational hyperspectral imaging.
Currently, data-driven methods, which are mainly based on deep learning \cite{NUAAACM,AGPCNet,RDIAN}, have become the mainstream schemes to learn feature representations and target detection inference from labeled dataset samples \cite{chen2024micpl}.
%In view of this, in this work we mainly focus on the challenges faced by data-driven methods.
}
% In contrast, data-driven methods have become a type of mainstream schemes in recent years \cite{NUAAACM}. They can effectively learn target detection tasks from labeled datasets \cite{AGPCNet,RDIAN},  

Furthermore, according to the {difference} of {image-frame numbers}, existing detection methods could be further divided into single-frame and multi-frame schemes \cite{du2021spatial}.
Generally, single-frame ISTD (SISTD) {methods, e.g., DNANet \cite{DNAnetli2022dense}, RISTD \cite{hou2021ristdnet} and RPCANet \cite{wu2024rpcanet},} extract target features only from spatial domain, {often with} obvious advantages of low complexity and high detection speed.
% Generally, single-frame ISTD (SISTD) does not need to consider the relations between multiple consecutive frames \cite{DNAnetli2022dense}, as in \cite{hou2021ristdnet} and \cite{wu2024rpcanet}. These methods extract target features only from the spatial domain of images \cite{ISNetzhang2022isnet}, owning some obvious advantages of low complexity and high detection speed.
%However, \red{this type of detection} methods are \red{often} not always effective in complex scenes (e.g., obscured by obstacles or overwhelmed by clutter and noise). 
% The multi-frame ISTD (MISTD) could improve detection performance by extracting temporal context from sequential frames to enhance target representation, as shown in Fig. \ref{fig:intro}(a).
{In contrast}, multi-frame ISTD (MISTD) methods could further improve detection performance. {This improvement is attributed to that these methods, {e.g., STDMANet \cite{yan2023stdmanet}, ST-Trans \cite{tong2024strans} and SSTNet \cite{chen2024sstnet}, can extract more discriminative} target features from both spatial and temporal domains, as shown in Fig. \ref{fig:intro}(a).}
% For example, STDMANet \cite{yan2023stdmanet} designs a feature extractor to obtain spatial-temporal features from multi-frame consecutive images,
%, and a spatial-temporal feature-based detection framework is proposed in \cite{du2021spatial}. 
% and SSTNet \cite{chen2024sstnet} employs a cross-slice ConvLSTM to leverage spatial-temporal contexts also on multiple frames. 
%\red{They both are exactly based on the target features from spatial and temporal domains.}

% 写当前多帧方法的不足，忽略了频域
Currently, almost all MISTD methods \cite{yan2023stdmanet, tong2024strans,chen2024sstnet} are {only} based on spatio-temporal {features}. 
{This type of approaches could achieve good detection performance. Nevertheless, they often fail to sufficiently capture the intricate characteristics of moving small targets. %Specifically, they often fail to capture the intricate characteristics of small moving targets entirely.
Besides, some targets in complex backgrounds could usually not exhibit a high correlation in consecutive frames.
These limitations above make it challenging to precisely discern infrared target motion details and contours, ultimately decreasing detection performance.
}
Generally, the information from different domains is complementary \cite{MaxMeandeshpande1999max}. This could help to characterize infrared small target features more comprehensively. 
%In view of this, \red{it has been widely explored in multi-modal information field.} 
{Therefore, our initial motivation is to introduce more information domains for infrared feature learning, further strengthening infrared small target detection paradigm.}

% At present, the existing schemes of MISTD are almost based on the spatial-temporal domain, and almost no methods consider the frequency information in images.nearly
% Currently, almost all MISTD methods are based on the spatial-temporal domain.
% with few considering the frequency information in images.
% They precisely capture the spatial-temporal contexts to reduce interference or enhance target features. 
% to solve the problem of small target sizes. 
% Nevertheless, these spatial-temporal domain-based methods could suffer from information loss due to blurring caused by the fast motion of targets or cameras. 
%  
% Nevertheless, while these approaches are promising, they often fail to capture the intricate characteristics of small moving targets entirely.
% Besides, targets in intricate backgrounds could not exhibit a high correlation in consecutive frames.
% , especially when the background changes. 
% due to camera motion. 
% In these challenging situations, obtaining targets' motion details and contours is extremely difficult, decreasing the detection performance.
% The information from different domains is complementary, which helps to characterize the target features more comprehensively.
% Therefore, it is compelling to introduce more information domains in feature learning. 

% Consequently, a vital issue for MISTD is capturing fine-grained target features without being affected by complex scenarios. 
Many image processing studies, {such as} \cite{li2015fre} and \cite{xu2020frelearn}, have proved that frequency domain features can provide abundant frequencies information.
{
Meanwhile, more researches have demonstrated that frequency information is significantly practical in image feature modeling, e.g., DAWN \cite{1145}, FMRNet \cite{1609} and frequency-assisted mamba \cite{4964}}.
% For example, \cite{1145} employs Wavelet transform for image deraining. \cite{1609} proposes a frequency mutual revision network to harness the strengths of both spatial and frequency domains. \cite{4964} introduces a frequency-assisted mamba for remote sensing image super-resolution.
{Besides, observed in frequency domain, noise usually distributes at high frequencies, while targets distribute at low frequencies \cite{chen2013lcm}.
Based on these findings, frequency information is believed to be potentially valuable in infrared small target detection.}
% Frequency domain processing could reduce noise disturbance, effectively improving the accuracy of ISTD. 
%\red{Therefore, on traditional spatio-temporal domains}, we develop a new triple-domain feature learning scheme \red{by} integrating frequency domain, as shown in Fig. \ref{fig:intro}(b).
%However, further exploration is needed to effectively acquire and integrate features from both spatial-temporal and frequency domains simultaneously.

To further enhance {infrared} target feature learning, {on traditional spatio-temporal domains}, we propose a new \emph{Triple-domain Strategy} (Tridos) {with} frequency-aware memory enhancement, as shown in Fig. \ref{fig:intro}(b).  
{Our scheme} is dedicated to exploiting the potential of frequency information {and enhancing infrared} target representation comprehensively. 
% to provide a more comprehensive target characterization.  
% Our scheme concentrates on the targets' characterization in the spatial-temporal domain and their expressions and variations in the frequency domain.
% Our scheme concentrates on depicting targets in the spatial-temporal domain and capturing their dynamic variations in the frequency domain for more accurate detection.
{In details,} our {work focuses} on depicting infrared targets simultaneously in spatio-temporal-frequency domains, {specially leveraging} their dynamic variations in frequency domain {to} promote target detection performance.
{Our Tridos is not a simple fusion of frequency and spatio-temporal feature learning. By contrast, it captures frequency features and then enhances spatio-temporal features using frequency features, even concerning cross-domain mismatches.
It is believed to be the first work to model the MISTD task from a frequency-aware spatio-temporal perspective.}
% It is the first time to model the task of MISTD from a frequency-aware spatial-temporal perspective. 
In summary, the main contributions {of our work} are summarized as follows:

% (I) Unlike previous methods, we propose a new MISTD scheme to extract target features across dual domains. Our Tridos could effectively resolve the problems of motion blur and clutter interference, which enhances detection robustness.
% systematically employ both spatial-temporal and frequency features. 
% This scheme effectively resolves the problem of motion blur and clutter interference, which enhances detection robustness.
(I) We explore and propose Tridos, a pioneering triple-domain scheme to extend the feature learning perspective for MISTD. 
{In addition to} traditional spatio-temporal domains, it can capture target features from frequency domain and then realize the fusion and enhancement of spatio-temporal-frequency features.

(II) Based on Fourier transform, a local-global frequency-aware module is developed to extract comprehensive frequency features from local and global perceptual patterns. 
{Additionally}, inspired by human visual system, a memory-enhanced spatial relationship module is designed to model the inter-frame correlations of small targets.   

(III) A residual compensation unit is constructed to eliminate the possible feature mismatches between different domains, {and} then auxiliarily fuse \& enhance spatio-temporal-frequency features of small targets. 
% different domain features. 

(IV) 
%By {re-defining} traditional regression loss,
A new dual-view regression loss function is re-defined to optimize model training, especially {tailored} for {ISTD}. 

%We further introduce a dual view regression loss to extend traditional IoU loss, which is more suitable for detecting infrared small targets.

% (IV) Referencing available repositories, we develop a new multi-frame infrared small target dataset (ITSDT-15K) with accurate bounding box annotations to advance the field.

\section{Related Work}
\subsection{Single-frame Infrared Small Target Detection}
Single-frame infrared small target detection deals with stationary targets in a single image. {It} can {often} be divided into model-driven and data-driven methods. 

Model-driven methods {detect small} targets by exploiting image characteristics for target enhancement or background suppression, {often with} threshold segmentation \cite{chen2023augtarget}. {This types of methods} can be further categorized into filter-based, human visual system (HVS)-based, and data structure-based methods.
Filter-based methods utilize the difference between target and background pixels to highlight targets and remove background noise interference, {e.g.,} Top-hat \cite{Tophat} and Maxmean \cite{MaxMeandeshpande1999max}.
HVS-based methods extract salient regions by measuring the maximum contrast between center pixel and {its} neighboring regions, e.g., local contrast measure (LCM) \cite{chen2013lcm} and its modified version{s}: RLCM \cite{han2018infraredrlcm}, HBMLCM \cite{han2019hbmlcm} and WSLCM \cite{han2020wslcm}.
Data structure-based methods employ the {special} structure of infrared images to separate {targets from background}, e.g., 
%IPI \cite{gao2013infraredIPI} and 
SMSL \cite{wang2017SMSL}. 
Although these model-driven approaches have achieved outstanding performance, they {could often not} adapt to intricate environments.
% Although these traditional approaches have achieved excellent performance in specific scenarios, their effectiveness relies heavily on 
% image processing techniques and manually designed features. It limits their ability to adapt to intricate environments.
% Recently, some methods based on data structure have been promoted, such as IPI \cite{gao2013infraredIPI} and SMSL \cite{wang2017SMSL}.  
% This scheme mainly utilizes the structure of infrared images to separate the background with nonlocal self-similarity and the target with sparse properties.

{In contrast}, data-driven methods primarily exploit deep neural networks that can adaptively learn target features by {the} training on {numerous} labeled samples. {These ones have} become dominant, with the availability of infrared small target detection datasets \cite{NUAAACM,DNAnetli2022dense,ISNetzhang2022isnet}.
For example, ACMNet \cite{NUAAACM} integrates low-level details and high-level semantics through asymmetric contextual modulation, and {it} further introduces dilated local contrast measurements in ALCNet \cite{dai2021ALCNet}. 
DNANet \cite{DNAnetli2022dense} designs a dense nested interactive network to enhance the features of small targets.
Moreover, AGPCNet \cite{AGPCNet} develops an attention-guided pyramid context network to obtain a global association of semantics.
RDIAN \cite{RDIAN} uses multi-scale convolutions to capture diverse target features and extend receptive fields.
UIUNet \cite{wu2022uiuNet} integrates two U-nets to learn multi-sale and multi-level {feature} representations. 
ISNet \cite{ISNetzhang2022isnet} creatively emphasizes the importance of target shapes.
% IR-TransDet \cite{IRTransDet} combines convNets with transformer to extract global semantic information.
Furthermore, EFLNet \cite{yang2024eflnet} constructs a feature-enhancing learning network, and
RPCANet \cite{wu2024rpcanet} proposes an interpretable deep neural network with theory-guided to replace matrix computation.

{Totally, } single-frame infrared small target detection has been widely studied and has performed well. However, this {type of} pipeline {cannot always} utilize target motion features. It is susceptible to noise and background interference, especially in challenging {detection} scenarios.

\subsection{Multi-frame Infrared Small Target Detection}
In contrast, multi-frame {detection} schemes {usually} process multiple consecutive frames simultaneously.
Therefore, {how to extract} the spatio-temporal features of targets has received extensive attention.

In {representative} traditional schemes, optical flow-based methods \cite{optical, kwan2020optical1} utilize {the} brightness variations in different frames to describe target motion features. 
STCP \cite{optical} employs optical flow algorithm to compute the dense trajectory of targets, {then it creates} a binary image to extract salient contours as candidate target regions.
Some methods construct spatio-temporal tensors to distinguish targets from backgrounds, such as 4D STT \cite{wu20234Dtensor} and NPSTT \cite{wang2021NPSTT}. 
To solve the issue of inaccurate background estimation, NTLA \cite{liu2021nonconvex} proposes a nonconvex tensor low-order approximation strategy.
FST-FLNN \cite{luo2024feedback} improves tensor nuclear norm {by} log operation for target enhancement and background suppression. 
Moreover, energy accumulation-based methods \cite{zhang2005enery, ren2019energy2} can effectively improve SNR and enhance the energy of small targets.
{D}ynamic programming-based methods \cite{sun2017dp} search for the optimal motion trajectory of targets {usually} through dynamic programming algorithm.
Although these traditional methods have made some progress, they rely on prior knowledge heavily.

To overcome the problems {appearing in} traditional methods, some {new} approaches apply deep learning to MISTD. 
This {type of} pipeline can exploit the temporal association between sequential frames to improve {target} detection performance. 
However, it {remains largely} {un}explored due to the lack of adequate MISTD datasets with high-quality annotations. 
{In recent years, some} representative methods have {been proposed}. For example, the inter-frame energy accumulation enhancement is proposed in \cite{du2021spatial}.  
STDMANet \cite{yan2023stdmanet} introduces a spatio-temporal differential multiscale attention network, a {pioneering} study {with} temporal attention mechanism.
DTUM \cite{li2023directiondtum} is {designed} to encode motion direction and extract {target} motion information. 
% % Meanwhile, it presents a multi-frame infrared dim target dataset (NUDT-MIRSDT) for segmentation. 
Moreover, SSTNet \cite{chen2024sstnet} {devises} a sliced spatio-temporal network to employ cross-slice motion context.
Recently, ST-Trans \cite{tong2024st-trans} introduces a spatio-temporal transformer to learn {inter-frame associations and} handle complex {detection} scenarios. 
However, these methods {solely focus on} effectively extracting and {utilizing} target features {only} from spatio-temporal domains, {neglecting} the utilization of frequency information implied in consecutive frames.

Therefore, in this work, we focus on {introducing frequency-domain information to expand target feature learning.}
% applying the frequency-domain information of sequential frames to traditional spatial-temporal networks, enhancing target feature learning.
{For intuitiveness and clarity, we summarize the typical differences between our Tridos and related works, as follows:}

{(I) Triple-Domain Feature Learning: existing works focus on feature learning from spatio-temporal domains. In contrast, our method leverages additional frequency domain features to refine the detection paradigm.} 

{(II) Inter-Frame Relation Mining: previous methods usually compute the weights of each frame, and only consider simple inter-frame relationships. Different from them, our method mimics human memory mechanisms to capture the global dependencies of targets in different time steps.}

{(III) Cross-Domain Feature Reconciling: unlike existing works that use some classic operations such as convolution and concatenation for feature fusion, we use residual compensation to ameliorate cross-domain feature collaboration.}

% the detection performance of MISTD. 
% multi-frame infrared small target detection.
% for infrared time-sensitive target detection and tracking \cite{ITSDT}.
%Furthermore, referencing available repositories, we construct a new MISTD dataset (ITSDT-15K) with {precise} bounding box annotations to {further} validate our proposed method {for} MISTD. 
%facilitate the development of  Their details are in section \ref{experiments}.
% Referencing available repositories, we develop a new multi-frame infrared small target dataset (ITSDT-15K) with accurate bounding box annotations to advance the field.
\section{METHODOLOGY}
\subsection{Overall Architecture}

\begin{figure*}[t]
    \centering
\includegraphics[width=\linewidth]{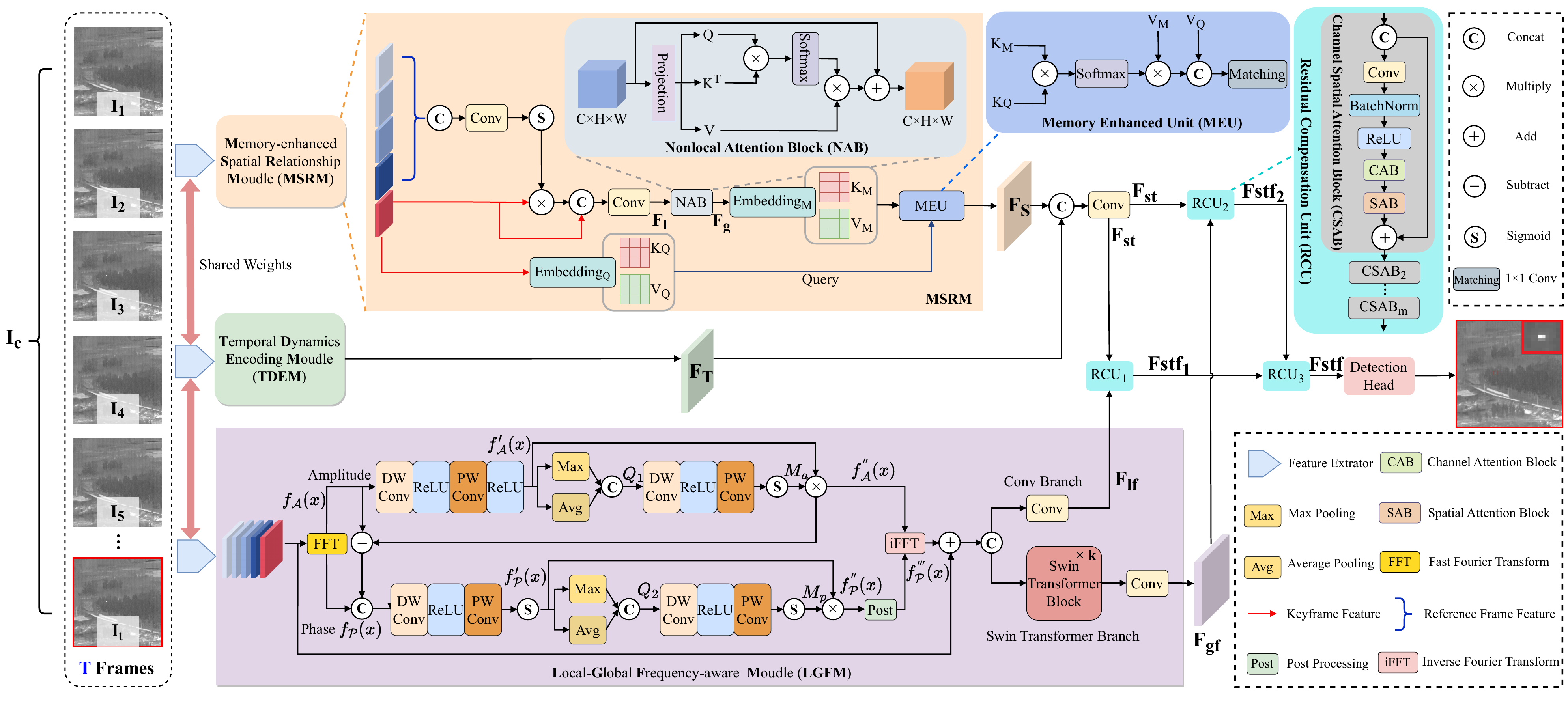}
    %\caption{Overview of proposed framework Tridos. $\textbf{I}_\textbf{c}$ is a collection of consecutive frames randomly sampled from infrared video. \red{The backbone with shared weights extracts target features from these frames. Then, the extracted features are processed through three key modules. (1) Memory-enhanced Spatial Relationship Module (MSRM): Captures spatial relationships between frames to enhance feature representation. (2) Temporal Dynamic Encoding Module (TDEM): Encodes temporal dynamics features via differential learning and residual enhancing. (3) Local-Global Frequency-aware Module (LGFM): Extracts frequency features using Fourier transform to enhance target representation in the frequency domain. The features from these three branches are then reconciled using Residual Compensation Units (RCUs), which address potential feature mismatches between different domains. Finally, the compensated features are passed through the detection head to produce the final detection results.}}
     \caption{Overview of proposed framework Tridos. $\textbf{I}_\textbf{c}$ is a group of video frames for Tridos. 
     {It consists of a backbone and three primary branches, i.e., a} \emph{Memory-enhanced Spatial Relationship} extraction branch, {a} \emph{Temporal Dynamic Encoding} branch, {and a} \emph{Local-Global Frequency-aware} branch. {The features extracted by the first two branches are fused together to generate} $\boldsymbol{F_{st}}$. The {two outputs of the third} branch, $\boldsymbol{F_{lf}}$ and $\boldsymbol{F_{gf}}$ {are fused with} $\boldsymbol{F_{st}}$ {by RCU to generate} $\boldsymbol{F_{stf_1}}$ and $\boldsymbol{F_{stf_2}}$, respectively. {Finally,} $\boldsymbol{F_{stf_1}}$ and $\boldsymbol{F_{stf_2}}$ are further compensated by RCU to obtain the {refined features} $\boldsymbol{F_{st}}$ for designed detection head.}
    \label{fig:frame}
\end{figure*}

\begin{figure}[t]
    \centering
\includegraphics[width=\linewidth]{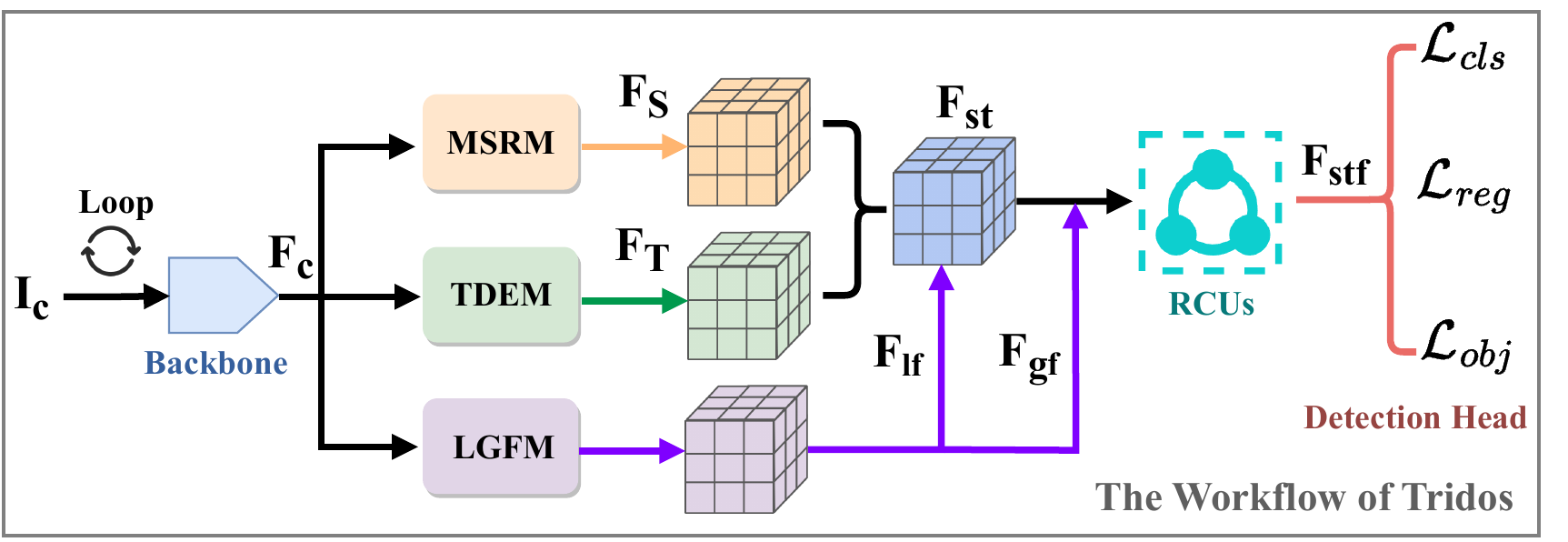}
    \caption{The workflow of Tridos. It contains the calculation process of input $\textbf{I}_\textbf{c}$ and the feature flow of target detection.
    }
    \label{fig:workflow}
\end{figure}

We aim to {effectively promote the detection performance of MISTD by triple-domain feature learning.} {A new framework is proposed to fulfill this goal, as presented in Fig. \ref{fig:frame}. For clarity, its simplified workflow is shown in Fig. \ref{fig:workflow}.}
% The overall workflow of our Tridos is presented in Fig. \ref{fig:workflow}. 

% accept
In our {framework, it} randomly sample a collection of consecutive frames $\textbf{I}_\textbf{c} = \{\textbf{I}_\textbf{1}, \textbf{I}_\textbf{2}, \cdots, \textbf{I}_\textbf{t} \}$ from {an} infrared video with time window size of $T = t$ as input. $\boldsymbol{I_t}$ is the keyframe that needs to be detected, and the other ones are reference frames {used for} provid{ing} context information.
Then, we feed each frame into {the} feature extractors with shared weights to obtain multi-frame features $\boldsymbol{F_c} = \{\boldsymbol{F_1}, \boldsymbol{F_2}, \cdots, \boldsymbol{F_t} \} \in \mathbb{R}^{T \times C \times H \times W}$, where $C$, $H$ and $W$ indicate the channel, height and width of feature matrix, respectively. Specifically, we adopt {the} CSPDarknet \cite{ge2021yolox} 
%pre-trained on the COCO \cite{lin2014COCO} dataset 
as backbone. 
% Furthermore, the feature extractors share weights. 
% We design three branches to construct frequency-aware spatial temporal networks through MSRM, TDEM, and LGFM.
We design three branches to realize the strategy of  triple-domain feature learning through memory-enhanced spatial relationship module (MSRM), temporal dynamic encoding module (TDEM), and local-global frequency-aware module (LGFM).
The residual compensation unit (RCU) {then} aims to eliminate the {potential} feature mismatches across different domains while fusing and enhancing different domain features.
% achieve interaction between frequency information acquired by LGFM and the spatio-temporal features. 
{Afterwards}, the adequately interacted spatio-temporal-frequency features, $\boldsymbol{F_{stf}}$ are delivered to {a} detection head to obtain final {detection} results.

MSRM receives multi-frame features $\boldsymbol{F_c}$ as {its} input and generates the memory-enhanced spatial features $\boldsymbol{F_S}$.
Similarly, we introduce TDEM to extract target motion features $\boldsymbol{F_T}$ by employing differential learning and residual enhancing between neighboring frames.
{To incorporate} both the location and motion information of targets, we concatenate spatial features $\boldsymbol{F_S}$ and temporal features $\boldsymbol{F_T}$ to {form} spatio-temporal features $\boldsymbol{F_{st}}$, as follows:
\begin{equation}
    \boldsymbol{F_{st}} = Conv(Concat\left[\boldsymbol{F_S}, \boldsymbol{F_T}\right])
\end{equation}
where $Conv$ denotes a $3\times3$ convolutional layer, {with} batch normalization and SiLU activation function. 
Inspired by Fourier Transform, we transform multi-frame features $\boldsymbol{F_c}$ into frequency domain. LGFM is proposed for modeling the local-global variation of targets in frequency domain, enhancing target representation {from} different {perspectives.}
\begin{equation}
    \boldsymbol{F_{lf}, F_{gf}} = LGFM(\boldsymbol{F_c})
\end{equation}
where $\boldsymbol{F_{lf}}$ and $\boldsymbol{F_{gf}}$ {indicate the} local and global features in frequency domain, respectively.
Then, $\boldsymbol{F_{lf}}$, $\boldsymbol{F_{gf}}$ and $\boldsymbol{F_{st}}$ are fed into the designed RCUs to {facilitate} feature interaction and alleviate the {feature} mismatches in different domains. {Their} calculation processes {are} as follows:
\begin{equation} \label{eqRCU}
\left\{
\begin{split}
\boldsymbol{F_{stf_1}} &= RCU_1(\boldsymbol{F_{lf}},\boldsymbol{F_{st}})
\\ \boldsymbol{F_{stf_2}} &= RCU_2(\boldsymbol{F_{gf}},\boldsymbol{F_{st}})
\end{split}
\right.
\end{equation}
% \begin{equation} \label{eqRCU}
% \left\{
% \begin{split}
% \boldsymbol{F_{stf_1}} &= RCU_1(\boldsymbol{F_{lf}},\boldsymbol{F_{st}})
% \\ \boldsymbol{F_{stf_2}} &= RCU_2(\boldsymbol{F_{gf}},\boldsymbol{F_{st}})
% \\ \boldsymbol{F_{stf}} &= RCU_3(\boldsymbol{F_{stf_1}},\boldsymbol{F_{stf_2}})
% \end{split}
% \right.
% \end{equation}
After that, we can obtain the compensated spatio-temporal-frequency features ${\boldsymbol{F_{stf}} = RCU_3(\boldsymbol{F_{stf_1}},\boldsymbol{F_{stf_2}})}$.
Finally, {infrared target} detection results are acquired by the decoupled head of YOLOX \cite{ge2021yolox} and non-maximal suppression.
% How to integrate the features of different domains is a challenging problem. Previous methods fuse spatial and temporal features via concatenation \cite{du2021spatial} or multilayer convolution \cite{yan2023stdmanet}. They seldom consider the problem of misalignment in different domain features.

\subsection{Memory-enhanced Spatial Relationship Module}
Spatial relationship modeling is {crucial} for accurate{ly predicting and tracking target locations.} Previous methods directly {combine} multi-frame spatial features via concatenation \cite{du2021spatial} or multi-layer convolution \cite{yan2023stdmanet}. 
{However}, {these ones} seldom consider the keyframe's global context information and the spatial dependencies between neighboring frames. 
To {address} this, we propose MSRM, as shown in Fig. \ref{fig:frame}.
In {it}, non-local attention block (NAB) is introduced to {capture} the long-distance dependencies between pixels in different time steps. 
When tracking moving targets, humans can recognize targets more accurately and extract target's spatial relationship{s} by comparing and inferring the features between {a} keyframe {and} its reference frames.
Therefore, inspired by human visual system, we devise the memory enhancement unit (MEU) to expand target features. 
It could fully employ multi-frame features by creating key-value pairs to better capture {the} spatial relationships between targets.

Specifically, in MSRM, we first {merge the features of} reference frames and keyframe through following computational process. We assign different weights to different pixels {in} reference frames.
\begin{align}
\begin{split}
\boldsymbol{\hat{F_{t}}} &= \sigma ( Conv(Concat[\boldsymbol{F_{1}},\cdots,\boldsymbol{F_{t-1}}])) \otimes \boldsymbol{F_t} \\ 
\boldsymbol{F_l} &= Conv(Concat[\boldsymbol{\hat{F_{t}}}, \boldsymbol{F_t}])
\end{split}
\end{align}
where $\sigma$ denotes {a} Sigmoid function, $\otimes$ is {element-wise} multiplication, $\boldsymbol{\hat{F_{t}}}$ {denotes} updated keyframe feature, and $\boldsymbol{F_l}$ indicates local inter-frame spatial relations.

Then, we send $\boldsymbol{F_l}$ to NAB to explore the correlation {between} different pixels {across} frames, {overcoming} the local {limitations of} convolutional layers, as follows:
\begin{equation}
    \begin{split}
        \boldsymbol{Q},\boldsymbol{K},\boldsymbol{V} &= Projection(\boldsymbol{F_l})\\
         \boldsymbol{F_g} &= \gamma \cdot Softmax(\frac{\boldsymbol{Q}\boldsymbol{K^T}}{\sqrt{d}})V + \boldsymbol{F_l}
    \end{split}
\end{equation}
where $Projection(\cdot)$ is a linear transform implemented by $1\times1$ convolution, $\gamma$ is a hyper-parameter, $\sqrt{d}$ is the scale factor for normalization, and $\boldsymbol{F_g}$ denotes inter-frame local-global spatial relations. 
In this way, any two pixels can interact and {be captured} long-distance dependencies, not just {being} limited to local neighboring {pixels}.

After that, we {store} the local-global spatial features $\boldsymbol{F_g}$ in memory and the keyframe features $\boldsymbol{F_t}$ as query, similar to the {visual} tracking mechanism of human brain.
In detail, we employ four distinct embedding layers to generate corresponding key-value pairs, {as follows}.
\begin{equation}
    \left\{\begin{split}
       \boldsymbol{K_Q}, \boldsymbol{V_Q} &= f_Q(\boldsymbol{F_t}) \\
       \boldsymbol{K_M}, \boldsymbol{V_M} &= f_M(\boldsymbol{F_g})
    \end{split} \right.
\end{equation}
where $f_Q(\cdot)$ and $f_M(\cdot)$ {represent} different embedding layers.
{\em K} is utilized to align the location information of targets, and {\em V} holds the high-level semantic features {of targets}.
In MEU, we first calculate the similarity {between} $\boldsymbol{K_M}$ and $\boldsymbol{K_Q}$, {with} $Softmax$ normaliz{ing}. Subsequently, the regions related to target features are queried from $\boldsymbol{V_M}$ and {then} concatenated with $\boldsymbol{V_Q}$. This establishes the connection between keyframe and {its} reference frames, {leading to the} improvement of target features with the visual information from memory. MEU is computed mathematically as follows:
\begin{equation}
    \begin{split}
         \boldsymbol{M_s} &= Softmax(\boldsymbol{K_M} 
 \otimes \boldsymbol{K_Q}) \\
  \boldsymbol{F_S} &= Matching(Concat[\boldsymbol{M_s} \otimes \boldsymbol{V_M}, \boldsymbol{V_Q}])
    \end{split}
\end{equation}
where $\boldsymbol{M_s}$ is the similarity matrix {of} $\boldsymbol{K_M}$ and $\boldsymbol{K_Q}$, $Matching$ represents $1\times1$ convolution, and $\boldsymbol{F_S}$ denotes the local-global spatial features by memory enhancement.

\subsection{Temporal Dynamics Encoding Module}
\begin{figure}[h]
    \centering
\includegraphics[width=0.7\linewidth]{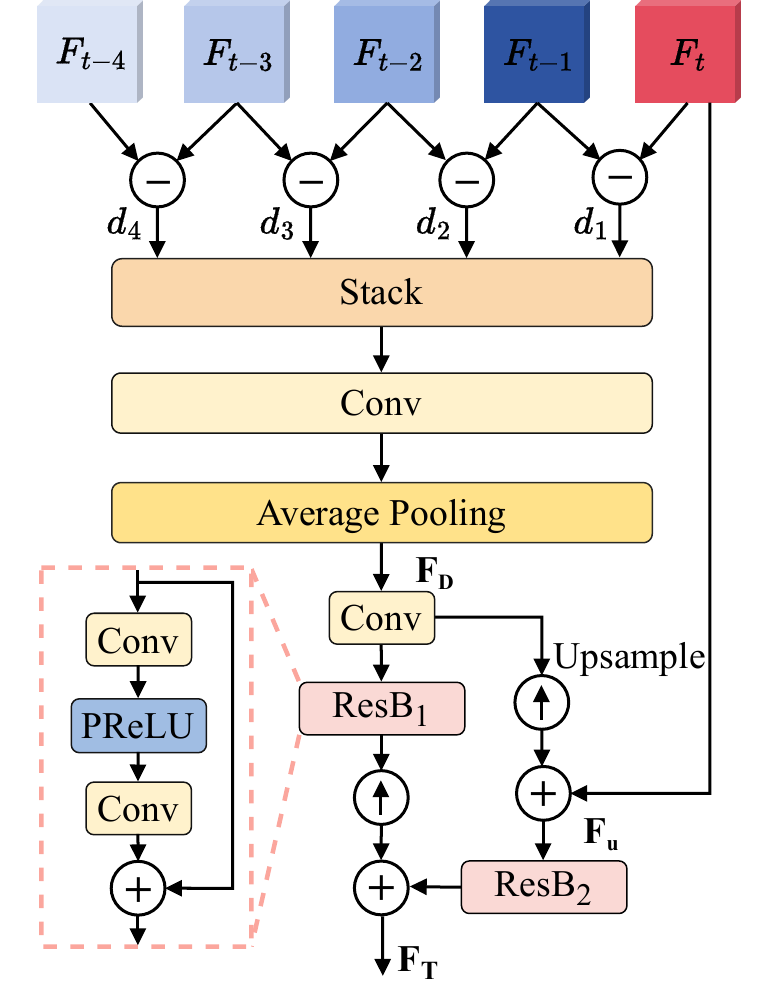}
    \caption{The details of our proposed TDEM, with time window $T = 5$. 
    $\text{ResB}_1$ and $\text{ResB}_2$ are two residual blocks for enhancing differential information.
    }
    \label{fig:tdem}
\end{figure}

How to extract the motion paradigm of targets is a challenging problem in MISTD. Previous methods {often} use optical flow \cite{kwan2020optical1} or recurrent neural networks \cite{chen2024sstnet}. However, since targets {are} too {small}, the motion information between neighboring frames is not {obvious, making {the extraction of} motion features challenging. Recently, temporal difference learning has emerged as an effective paradigm for extracting motion information from video sequences, as {in} LGTD \cite{xiao2023stdm} and TDN \cite{wang2021tdn}.
% For example, LGTD \cite{xiao2023stdm} systematically proposes a local-global temporal difference learning framework for video super-resolution. TDN \cite{wang2021tdn} devises a temporal difference network by explicitly leveraging a temporal difference operator for video action recognition. 
These works lay the foundation for utilizing temporal difference learning in various applications.} 
{Inspired by these}, we design a TDEM, as shown in Fig. \ref{fig:tdem}. It stacks all adjacent frames' differential information within time window {to solve the sparsity issue observed in a single difference.} 
{Unlike previous methods, we also incorporate residual blocks to amplify small targets' motion features through enhancing differential information.}
% We further introduce residual blocks for differential information enhancement to obtain more representative motion features.
In detail, TDEM dynamically encodes difference maps and propagates local motions into keyframe features. As shown in Fig. \ref{fig:tdem}, we set the size of time window {to} $T = 5$. Initially, we acquire continuous local motions $\boldsymbol{D} = [d_1,d_2,d_3,d_4]$ in sequential images by subtracting adjacent frames. 
Then, we use a $3\times3$ convolution and downsample through {an} average pooling with a stride of $2$ to extract the motion dependencies of targets $\boldsymbol{F_D}$. {This step} can be expressed as follows:
\begin{equation}
    \boldsymbol{F_D} = Avg(Conv(Stack(d_1,d_2,d_3,d_4)))
\end{equation}
where $Avg$ means average pooling and $Stack$ {represents a} splice operation on dimension $1$.
Downsampling {is executed} due to two main considerations. One is that differential information is sparse, and motions are more prominent in low-resolution space. {The other} is that pooling can reduce the computational cost of temporal dynamic encoding.

Moreover, we integrate {both} motion and keyframe features to cope with the dynamic changes of targets in temporal dimension. The motion dependencies, $\boldsymbol{F_D}$ {are restored as its} original size and then added to keyframe features. Furthermore, we {utilize} two residual blocks to {enhance} local {motion}, and deeply incorporate motion dependencies and target features. {They} can be calculated as follows: 
\begin{equation}
\begin{split}
      \boldsymbol{F_u} &= \boldsymbol{F_t} + f_u(Conv(\boldsymbol{F_D})) \\
      \boldsymbol{F_T} &= ResB_2(\boldsymbol{F_u}) + f_u(ResB_1(Conv(\boldsymbol{F_D})))
\end{split}
\end{equation}
where $f_u$ is {an}  upsample operation, {and} $\boldsymbol{F_T}$ denotes the motion features {obtained by} temporal dynamic encoding. $ResB$ represents the residual block which {consists of} two $3\times3$ convolutions and a PReLU activation function.

\subsection{Local-global Frequency-aware Module}
Frequency domain features {provide a comprehensive understanding to infrared small} targets at {various} frequencies, {while also reducing} image noise and interference. {Our objective is} to thoroughly explore the potential of frequency {for expanding} target feature learning.

As shown in Fig. \ref{fig:frame}, {designed} LGFM has two branches {dedicated to} model{ing} and extract{ing} frequency features from multiple views and levels.
One is the convolution branch, which captures local correlation{s} and spatial structure{s} from frequency. The other is the Swin Transformer \cite{2022videoswin} branch, which globally models and encodes frequency features {using} a sliding window mechanism. This branch improve{s} the receptive field, {allowing for} extract{ing} global contexts.

% given the input multi-frame features collection $\boldsymbol{F_c} \in \mathbb{R}^{T \times C \times H \times W}$, 
Concretely, the Fourier transform \cite{1982fft} $f_\mathcal{F}(\cdot)$ converts each frame's features {into} frequency domain. Simply, for given features $x$, this process can be formulated as follows :
\begin{equation}
    f_\mathcal{F}(x)(u,v)=\sum_{h=0}^{H-1}\sum_{w=0}^{W-1}x(h,w)e^{-j2\pi\left(\frac hHu+\frac wWv\right)}
\end{equation}
where $u$ and $v$ represent the horizontal and vertical components of frequency.
$H$ and $W$ are the height and width of {given} $x$.
The frequency domain information, $f_\mathcal{F}(x)$ can be further decomposed {into} 
amplitude $f_\mathcal{A}(x)$ and phase $f_\mathcal{P}(x)$, {by}  
\begin{equation}
    \left\{\begin{split}
         f_\mathcal{A}(x)(u,v) &=\begin{bmatrix}\mathcal{R}^2(x)(u,v)+\mathcal{I}^2(x)(u,v)\end{bmatrix}^{1/2} \\
         f_\mathcal{P}(x)(u,v) &=\arctan\left[\frac{\mathcal{I}(x)(u,v)}{\mathcal{R}(x)(u,v)}\right]
    \end{split}\right.
\end{equation}
where $\mathcal{R}(x)$ and $\mathcal{I}(x)$ {represent} the real and imaginary part of $f_\mathcal{F}(x)$, respectively.

The motivation for decomposing frequency features into amplitude and phase is to employ the information from these two aspects. Amplitude reflects the feature energy distribution {of} different frequencies, {while} phase comprises the position and relative relationship in frequency domain.
Initially, we process amplitude and then utilize amplitude residuals to further extract and {refine} phases.
{To achieve this}, we use a $3\times3$ depth-wise convolution, i.e., $DW(\cdot)$, to independently learn spatial context information on each channel, {and a} cross-channel blending with $1\times1$ point-wise convolution, i.e.,  $PW(\cdot)$, {to} aggregate pixels, as follows:
\begin{equation} \label{eqDP}
   f'_\mathcal{A}(x) = ReLU(PW(ReLU(DW(f_\mathcal{A}(x)))))
\end{equation}

To enhance the energy of moving {infrared small} targets, we apply max pooling $Max(\cdot)$ and average pooling $Avg(\cdot)$ {along} channel dimension and {then} concatenate them. {We apply} the {similar} transform{ation} {on} $f'_\mathcal{A}(x)$ to get the amplitude attention map $\boldsymbol{M_a}$. {Then}, $\boldsymbol{M_a}$ multipl{ies}  with $f'_\mathcal{A}(x)$ to get the final refined amplitude features $f^{''}_\mathcal{A}(x)$, which can be formulated as follows:
\begin{equation}
\left\{\begin{split}
\boldsymbol{Q_1} &= Concat[Avg(f'_\mathcal{A}(x)), Max(f'_\mathcal{A}(x))]\\
\boldsymbol{M_a} &= \sigma (PW(ReLU(DW(\boldsymbol{Q_1}))))\\
f^{''}_\mathcal{A}(x) &= f'_\mathcal{A}(x) \otimes \boldsymbol{M_a}
\end{split}\right.
\end{equation}
where $\boldsymbol{Q_1} \in \mathbb{R}^{2\times H \times W}$ and $\sigma$ is {a} $Sigmoid$ function.

After that, we employ amplitude residuals $f_\mathcal{R}(x) =f^{''}_\mathcal{A}(x) - f_\mathcal{A}(x)$, which contain subtle amplitude variations to modify input and guide phase features extraction. 
{Regarding} phases, we {follow} a similar procedure {to} amplitudes, {by}
\begin{equation}
    \left\{\begin{split}
         f'_\mathcal{P}(x)&=\sigma(PW(ReLU(DW(Concat[f_\mathcal{P}(x), f_\mathcal{R}(x)])))) \\
         \boldsymbol{Q_2} &= Concat[Avg(f'_\mathcal{P}(x)), Max(f'_\mathcal{P}(x))] \\
         \boldsymbol{M_p} &= \sigma (PW(ReLU(DW(\boldsymbol{Q_2}))))\\
f^{''}_\mathcal{P}(x) &= f'_\mathcal{P}(x) \otimes \boldsymbol{M_p}
    \end{split}\right.
\end{equation}

Moreover, we {apply} post-processing to {restrict} the phase values to $[-\pi,\pi]$, which maintain{s} the continuity and stability of phases. For combining amplitude and phase, we employ the $Cos(\cdot)$ and $Sin(\cdot)$ functions to compute the real and imaginary components {of features}, as follows:
\begin{equation}
    \left\{\begin{split}
         f^{'''}_\mathcal{P}(x) &= 2\pi \cdot f^{''}_\mathcal{P}(x) - \pi \\
         \mathcal{R}'(x)(u,v) &= f^{''}_\mathcal{A}(x)(u,v) \cdot Cos(f^{'''}_\mathcal{P}(x)(u,v))\\
         \mathcal{I}'(x)(u,v) &= f^{''}_\mathcal{A}(x)(u,v) \cdot Sin(f^{'''}_\mathcal{P}(x)(u,v))
    \end{split}\right.
\end{equation}
where $x'_\mathcal{F} = \mathcal{R}'(x) + \mathcal{I}'(x)j$ indicates integrated frequency features. Then, the inverse Fourier transformation $f^{-1}_\mathcal{F}(\cdot)$ is used to transform $x'_\mathcal{F}$ {back} to {its} original domain.
Suppose the input feature is $x$, {its} final output $x'$ is obtained by adding $x$ and {its} transformed features through residuals, {by}:
\begin{equation}
x' = x + f^{-1}_\mathcal{F}(x'_\mathcal{F})
\end{equation}
% where $f^{-1}_\mathcal{F}(\cdot)$ denotes the inverse Fourier transformation. 

For the input {of} multi-frame feature collection $\boldsymbol{F_c}$, we process each frame separately to {obtain} outputs $\boldsymbol{F'_c} = \{\boldsymbol{F'_1}, \boldsymbol{F'_2}, \cdots, \boldsymbol{F'_{t}}\} \in \mathbb{R}^{T \times C \times H \times W}$. 
Finally, we splice them to feed into two separate branches, {as follows:}
\begin{equation}
    \begin{array}{c}
        \boldsymbol{F_{lf}} = ConvB(Concat[\boldsymbol{F'_1}, \boldsymbol{F'_{2}}, \cdots, \boldsymbol{F'_{t}}])\\
        \boldsymbol{F_{gf}} = SwinB(Concat[\boldsymbol{F'_1}, \boldsymbol{F'_{2}}, \cdots, \boldsymbol{F'_{t}}])
    \end{array}
\end{equation}
where $\boldsymbol{F_{lf}}$ {is}  the local frequency features from convolution branch $ConvB(\cdot)$. This branch contains two $3\times3$ convolution {layers} with  stride 1, batch normalization and SiLU activation function.
$\boldsymbol{F_{gf}}$ {denotes} the global frequency features,  {acquired} from a Swin Transformer branch, i.e.,  $SwinB(\cdot)$.

\subsection{Residual Compensation Unit}
Our scheme thoroughly explores the spatial, temporal and frequency domain features {of infrared targets}. 
However, {feature} mismatches could {often} happen due to the learning {path} differences {of} different domain features.
Therefore, we need to further compensate features to retain valuable information in {all triple} domains.
With the help of RCU, the differences and commonalities between {different-domain} features can be fused and enhanced. {This step} helps to {strengthen} the understanding to {samples}, and reduce the sensitivity to interference.

As shown in Fig. \ref{fig:frame}, our RCU comprises several channel spatial attention blocks (CSAB). For example, for local frequency features $\boldsymbol{F_{lf}}$ and spatio-temporal features $\boldsymbol{F_{st}}$, we first splice them {along} channel dimension and use a $3\times3$ convolution {also} followed by batch normalization and ReLU (denoted as $Conv$) {to} further enhance target features. Then, we introduce a channel attention block (CAB) and a spatial attention block (SAB) to weight {the} features from different channels and spatial locations, adaptively {focusing} on target significant information. Furthermore, feature compensation is performed by {a} residual branch. The residual compensation between $\boldsymbol{F_{lf}}$ and $\boldsymbol{F_{st}}$, {$\boldsymbol{F_{stf_1}}$} can be {computed by}
\begin{equation}
\left\{\begin{split}
\boldsymbol{F_r} &= Concat[\boldsymbol{F_{lf}}, \boldsymbol{F_{st}}] \\
     CSAB &= f_{SAB}(f_{CAB}(Conv(\boldsymbol{F_r}))) + \boldsymbol{F_r} \\
     \boldsymbol{F_{stf_1}} &= CASB_m(CASB_{m-1}(\cdots CSAB_1(\boldsymbol{F_r})))
\end{split}\right.
\end{equation}
where $m$ is the number of CSAB. 
% Similarly, we perform the same operations compensation $\boldsymbol{F_{lf}}$ with $\boldsymbol{F_{st}}$ and $\boldsymbol{F_{stf_1}}$ with $\boldsymbol{F_{stf_2}}$ for the final refinement.
Similarly, {on} $\boldsymbol{F_{gf}}$ and $\boldsymbol{F_{st}}$, 
we perform the same operations {as on} $\boldsymbol{F_{lf}}$ and $\boldsymbol{F_{st}}$ to obtain $\boldsymbol{F_{stf_2}}$. {Also similarly, on both $\boldsymbol{F_{stf_1}}$ and $\boldsymbol{F_{stf_2}}$}, final {infrared small target feature} refinement $\boldsymbol{F_{stf}}$ is {generated}.

\subsection{Dual View Regression Loss}
Following {a} general paradigm of detection \cite{AGPCNet}, {a traditional} loss function can be expressed as
\begin{equation} \label{eqloss}
\mathcal{L}=\lambda_{reg}\mathcal{L}_{reg}+\lambda_{cls}\mathcal{L}_{cls}+\lambda_{obj}\mathcal{L}_{obj}
\end{equation}
where $\mathcal{L}_{reg}$ is a bounding box regression loss, {usually calculated by IoU loss.} $\mathcal{L}_{cls}$ is a classification loss and
$\mathcal{L}_{obj}$ is a target probability loss. $\lambda_{reg}$, $\lambda_{cls}$ and $\lambda_{obj}$ are three hyper-parameters to balance loss terms.

For {traditional} $\mathcal{L}_{cls}$ and $\mathcal{L}_{obj}$, we employ sigmoid focal loss \cite{lin2017focalloss} to solve the problem of positive and negative sampling imbalance.
%In terms of $\mathcal{L}_{reg}$, {we design a new regression loss $\mathcal{L}_{dvr}$}.
Due to small target size, popular IoU loss could be insufficient to capture {the} detailed information {of} target distribution. {Considerting this, a}
normalized Gaussian {W}asserstein distance (NWD) loss \cite{xu2022nwd} can {be adopted to} minimize the difference between target boxes by introducing Gaussian distribution. It could efficiently model {the} target boxes of different sizes and positions.

IoU loss assists {our} network {to} accurately {localize} small targets, while 
 NWD loss learns {their} distributional properties. Therefore, we design {a new} ${\mathcal{L}_{dvr}}$ {on} IoU and NWD loss{es} to {further enhance traditional $\mathcal{L}_{reg}$, as follows:}
{
\begin{equation} \label{eqab}
\left\{\begin{array}{l}
    \mathcal{L}_{iou} = 1 - IoU(B_P, B_G) \\
    \mathcal{L}_{nwd} = 1 -  exp(-\frac{\sqrt{\Vert N_P, N_G\Vert^2_2}}{C}) \\
    \mathcal{L}_{reg}= \mathcal{L}_{dvr} = \alpha \mathcal{L}_{iou} + \beta \mathcal{L}_{nwd} \\
\end{array}\right.
\end{equation}}
where $B_P$ is a predicted bounding box, {while} $B_G$ is {a labeled box}. $N_P$ and $N_G$ {denote} the Gaussian distribution {of} predicted boxes and {that of labeled boxes}, respectively. $C$ is a constant related to datasets. {Coefficients} $\alpha$ and $\beta$ are two hyper-parameters {to balance loss terms}.

\section{EXPERIMENTS} \label{experiments}
\subsection{Datasets and Evaluation Metrics}

{Usually}, dataset quality, sample quantity and scenario diversity {could} significantly impact the detection performance of data-driven methods. Currently, there are {a} few datasets \cite{RDIAN, DAUB} for MISTD. These ones are intended for aerial-based applications and aim at airborne, not {for} ground-based vehicle targets.
% To improve the quality, quantity, and diversity of infrared multi-frame datasets, we reorganize an additional MISTD dataset, namely ITSDT-15K. 
Thus, we re-organize an additional MISTD dataset on this public repository \cite{ITSDT}, i.e., ITSDT-15K, to {extend} the quality, quantity and diversity of infrared multi-frame datasets.
%{Its} raw infrared images are from this public repository \cite{ITSDT}. 
The {dim-small} targets in these images are moving vehicles captured by UAV infrared cameras.
{For ITSDT-15K, we have removed all wrong image sequences and revised all labeled bounding boxes to ensure that they accurately represent infrared targets. Then, we randomly selected 15K images, and split them into the 10K training images of 40 videos and the 5K test images of 20 videos. Besides, all frame-images have the same resolution of \(640 \times 480\).
}
% We remove some wrong image sequences and revise the bounding boxes. Finally, our ITSDT-15K dataset contains 15K images, including 10K training images of 40 videos and 5K test images of 20 videos, with an image resolution of $640 \times 480$.

{New} ITSDT-15K dataset contains a variety of complex scenarios, including grass, forests, obstacles, and ground clutter. 
Overall, our ITSDT-15K is adequate for evaluating the performance of different methods for MISTD.
% The ITSDT-15K dataset contains a variety of complex scenarios, including grass, forests, obstacles, ground clutter, etc. 
% Overall, the ITSDT-15K dataset is adequate to evaluate the performance of different methods for MISTD.
Accordingly, we employ ITSDT-15K and two public datasets, DAUB \cite{DAUB} and IRDST \cite{RDIAN}, for {method} evaluation in experiments. {All compared methods are evaluated on the revised ITSDT-15K dataset, as fair as possible.}
{On} DAUB and IRDST, we follow \cite{chen2024sstnet} to divide {their} training sets and test sets. 
% \red{Table 1 shows the details of three datasets.}
% \begin{table}[h]
% \centering
% \caption{The details of DAUB, ITSDT, and IRDST.}
% \label{tab:dataset}
% \resizebox{\linewidth}{!}{
% \begin{tabular}{c|cc|ccc|c}
% \hline
% \textbf{Dataset} &
%   \textbf{Target} &
%   \textbf{Resolution} &
%   \textbf{Frames} &
%   \textbf{Training Set} &
%   \textbf{Testing Set} &
%   \textbf{Scene Type} \\ \hline
% DAUB  & drone   & 256×256 & 13778 & 8983  & 4795 & Aerial-based \\ \hline
% IRDST &
% \begin{tabular}[c]{@{}c@{}c@{}}helicopter, \\ airplane, \\ drone\end{tabular} &
%   \begin{tabular}[c]{@{}c@{}}720×480, \\ 934×696\end{tabular} &
%   40656 &
%   20398 &
%   20258 &
%   Aerial-based \\ \hline
% ITSDT & vehicle & 640×480 & 15000 & 10000 & 5000 & Land-based \\ \hline 
% \end{tabular}}
% \end{table}
% Quantitative Table

In terms of evaluation metrics, 
following the common practice of target detection paradigm, we apply Precision (\%), Recall (\%), F1 score (\%), and \emph{Average Precision} (\%) (e.g., $\text{mAP}_{50}$, the average Precision with an IOU threshold 0.5). {These metrics} can be formulated by
% \begin{equation}
%     \begin{split}
%         Precision &= \frac{TP}{TP + FP} \\
%          Recall &= \frac{TP}{TP + FN} \\
%          F1 &= \frac{2 \times Precision \times Recall}{Precision + Recall} 
%     \end{split}
% \end{equation}
\begin{align}
\begin{split}
    \text{Precision} &= \frac{\text{TP}}{\text{TP}+\text{FP}}\\
\text{Recall} &=  \frac{\text{TP}}{\text{TP}+\text{FN}}\\
\text{F1} &= \frac{2\times \text{Precision} \times \text{Recall}}{\text{Precision}+\text{Recall}}
\end{split}
\end{align}
where TP, FP and FN {denote} the number of correct detection {targets} (True positive), false alarms (False positive) and missed detection {targets} (False negative), respectively. {As a comprehensive metric to assess detection methods}, F1 score {is computed by combining both} {Precision} and {Recall}.

\subsection{Implementation Details}
In implementation, time window size $T$ is set to $5$. For all methods, we reshape the resolution of infrared frames to $512 \times 512$, as fair as possible.
We train our Tridos for 100 epochs with batch size 4. 
{Initial} learning rate is 0.001 and {decreased} adaptively with training epochs. Adam is employed as an optimizer with {a} CosineAnnealingLR scheduler and momentum 0.937. We initialize model weights with {a usual} distribution. For hyper-parameters, we set $\lambda_{reg}$, $\lambda_{cls}$ and $\lambda_{obj}$ in Eq. (\ref{eqloss}) to 5, 1 and 1, respectively. The $\alpha$ and $\beta$ in Eq. (\ref{eqab}) are both 0.5.
During model testing, the IoU threshold for non-maximum suppression is {always} 0.65, and only {the} predicted boxes with a confidence greater than 0.001 are retained.
Regarding hardware, we conduct {all} experiments on two Nvidia GeForce 3090 GPUs.

%accept  MaxMean 
\begin{table}[t]
    \centering
    \caption{The parameter settings for model-driven methods.}
    \label{tab:modelsetting}
     % \rowcolors{1}{pink}{pink}
    \resizebox{\linewidth}{!}{
    \begin{tabular}{l | l}
    \hline
      \textbf{Method}    &  \textbf{Parameters} \\ \hline
      MaxMedian \cite{MaxMeandeshpande1999max} &  filter size: 5 \\
      TopHat \cite{Tophat}    &  structure size: $5\times 5$ \\
      RLCM \cite{han2018infraredrlcm}  & scale = 3, k1 = [2, 5, 9], k2 = [4, 9, 16] \\
      HBMLCM \cite{han2019hbmlcm} & scale size: [3, 5, 7, 9], external window size: $15\times 15$\\
      PSTNN \cite{wang2021NPSTT}& patch size: 40, slide step: 40, $\lambda_L = 0.7$\\
      WSLCM \cite{han2020wslcm} & gaussian kernel: $\frac{1}{16}\begin{bmatrix}1&2&1\\2&4&2\\1&2&1\end{bmatrix}$, scale size: [3, 5, 7, 9]  \\
      % TLLCM& gaussian kernel: $\begin{bmatrix}1&2&1\\2&4&2\\1&2&1\end{bmatrix}$  \\
     % ADDGD&  scale size: 3, 5, 7, 9\\
     % LEF&  $\alpha = 0.5$, $h = 0.2$ \\
     % LIG&  $k=0.2$, $N=11$\\
    \hline
    \end{tabular}}
\end{table}

\subsection{Comparisons With Other Methods}
We choose several representative methods for comparison, including model-driven and data-driven ones. 
{The parameter settings for model-driven methods are listed in Table \ref{tab:modelsetting}. For a fair comparison, we retrain all data-driven methods according to the basic settings specified in their original papers. All the details of training and testing are consistent with our method. For example, training period is uniformly 100 epochs and initial learning rate is always set to 0.001 for all compared methods.  
}
% For the model-driven methods, such as TopHat \cite{Tophat}, MaxMean \cite{MaxMeandeshpande1999max}, RLCM \cite{han2018infraredrlcm}, HBMLCM \cite{han2019hbmlcm}, PSTNN \cite{wang2021NPSTT} and WSLCM \cite{han2020wslcm}. For the data-driven methods, such as ACM \cite{NUAAACM}, RISTD \cite{hou2021ristdnet}, ISNet \cite{ISNetzhang2022isnet}, UIUNet \cite{wu2022uiuNet}, SANet \cite{zhu2023SAnet}, AGPCNet \cite{AGPCNet}, RDIAN \cite{RDIAN}, DNANet \cite{DNAnetli2022dense}, ResUnet \& DTUM \cite{li2023directiondtum}, SIRST5K \cite{lu2024sirst5k}, MSHNet \cite{liu2024MSH}, RPCANet \cite{wu2024rpcanet} and SSTNet \cite{chen2024sstnet}. 
{Besides}, since most of {chosen} methods are based on pixel-level segmentation, {for fairness}, we follow the paradigm of combined detector \cite{chen2024sstnet}.
% {that can use bounding box-based datasets}
\begin{table*}[h]
\centering
\caption{Quantitative comparison results of different state-of-the-art methods on three datasets. The best and second best results are highlighted in \red{red} and \blue{blue}, respectively.
}
\label{tab:quantitave}
\resizebox{\linewidth}{!}{
\begin{tabular}{c|c|c|cccc|cccc|cccc}
\hline
\multirow{2}{*}{\textbf{Scheme}} &
  \multirow{2}{*}{\textbf{Methods}} &
  \multirow{2}{*}{\textbf{Publication}} &
  \multicolumn{4}{c}{\textbf{DAUB}} &
  \multicolumn{4}{c}{\textbf{ITSDT-15K}} &
  \multicolumn{4}{c}{\textbf{IRDST}} \\ \cline{4-15}
 & & & $\textbf{mAP}_{\textbf{50}}$ & \textbf{Precision} & \textbf{Recall} & \textbf{F1} & 
 $\textbf{mAP}_{\textbf{50}}$ & \textbf{Precision} & \textbf{Recall} & \textbf{F1} & $\textbf{mAP}_{\textbf{50}}$ &
  \textbf{Precision} & \textbf{Recall} & \textbf{F1} \\ \hline
\multirow{6}{*}{Model-driven} &
  MaxMean \cite{MaxMeandeshpande1999max} & SPIE 1999 & 10.71 & 20.38 & 53.87 & 29.57 &
  0.87 & 10.85 & 8.74 & 9.68 & 0.01 & 0.28 & 1.48 &  0.47 \\ 
 &
TopHat \cite{Tophat} & IPT 2006 & 16.99 & 21.69 & 79.83 & 34.11 &
  11.61 &  27.21 &  43.07 &  33.35 &
  1.81 & 18.22 &  10.60 &  13.40 \\
 & RLCM \cite{han2018infraredrlcm} & IEEE TGRS 2013 &  0.02 &  0.27 &  5.21 &  0.51 &
  4.62 &  15.38 &  30.76 &  20.50 &
  1.58 &  16.28 &  9.70 &  12.16 \\
 &  HBMLCM \cite{han2019hbmlcm} &  IEEE GRSL 2019 &  3.90 &  23.96 &  16.52 &  19.56 &
  0.72 &  7.97 &  9.37 &  8.61 &
  1.16 &  29.14 &  4.66 &  8.03 \\
 &  PSTNN \cite{wang2021NPSTT} &  RS 2019 &  17.31 &  25.56 &  68.86 &  37.28 &
  7.99 &  22.98 &  35.21 &  27.81 &
  1.45 &  16.28 &  9.70 &  12.16 \\
 &  WSLCM \cite{han2020wslcm} &  SP 2020 &  1.37 &  11.88 &  11.57 &  11.73 &
  2.36 &  16.78 &  14.53 &  15.58 &
  1.69 &  20.87 &  8.70 &  12.28 \\ \hline
\multirow{14}{*}{Data-driven} &
  ACM \cite{NUAAACM} &  WACV 2021 &  64.02 &  70.96 &  91.30 &  79.86 &  55.38 &  78.37 &
  71.69 &  74.88 &  52.40 &  76.33 &  69.32 &  72.66 \\
 &  RISTD \cite{hou2021ristdnet} &  IEEE GRSL 2022 &  81.05 &  83.46 &  98.27 &  90.26 &
  60.47 &  85.49 &  71.60 &  77.93 &  66.57 &  84.70 &  79.63 &  82.08 \\
 &  ISNet \cite{ISNetzhang2022isnet} &  CVPR 2022 &  83.43 &  89.36 &  94.99 &  92.09 &  62.29 &  83.46 &
  75.32 &  79.18 &  59.78 &  80.24 &  75.08 &  77.58 \\
 &  UIUNet \cite{wu2022uiuNet} &  IEEE TIP 2022 &  86.41 &  94.46 &  92.03 &  93.23 &  65.15 &  84.07 &
  78.39 &  81.13 &  56.38 &  80.95 &  70.29 &  75.25 \\
 &  SANet \cite{zhu2023SAnet} &  ICASSP 2023 &  87.12 &  93.44 &  94.93 &  94.18 &
  62.17 &  87.78 &  71.23 &  78.64 &  64.54 &  84.29 & 77.02 &  80.49 \\
 &
  AGPCNet \cite{AGPCNet} &  IEEE TAES 2023 &  76.72 &  82.29 &  94.43 &  87.95 &  67.27 &
  \red{91.19} &  74.77 &  82.16 &  59.21 &  79.47 &  75.51 &  77.44 \\
 &  RDIAN \cite{RDIAN} &  IEEE TGRS 2023 &  84.92 &  88.20 &
  97.27 &  92.51 &  68.49 &  90.56 &  76.06 &  82.68 &  59.08 &  77.99 &  76.35 &  77.16 \\
 &  DNANet \cite{DNAnetli2022dense} &  IEEE TIP 2023 &  89.93 &  92.49 &  98.27 &  95.29 &  70.46 &  88.55 &
  80.73 &  84.46 &  63.61 &  82.92 &  77.48 &  80.11 \\
 &  DTUM \cite{li2023directiondtum} &  IEEE TNNLS 2023 &  85.86 &
  87.54 &  \red{99.79} &  93.26 &  67.97 &  77.95 &  \blue{88.28} &  82.79 &  71.48 &
  82.87 &  \blue{87.79} &  \blue{85.26} \\
 &
  SIRST5K \cite{lu2024sirst5k} &  IEEE TGRS 2024 &  93.31 &  97.78 &  96.93 &  97.35 &
  61.52 &  86.95 &  71.32 &  78.36 &  52.28 &  76.12 &  69.07 &  72.42 \\
 &  MSHNet \cite{liu2024MSH} &  CVPR 2024 &  85.97 &  93.13 &  93.12 &
  93.13 &  60.82 &  89.69 &  68.44 &  77.64 &  63.21 &  82.31 &  77.64 &  79.91 \\
 &  RPCANet \cite{wu2024rpcanet} &  WACV 2024 &  85.98 &  89.38 &
  97.56 &  93.29 & 62.28 &  81.46 &  77.10 &  79.22 &  56.50 &
  77.77 &  73.80 & 75.73 \\

  &  TMP \cite{zhu2024tmp} &  ESWA 2024 &  92.87 &  98.01 &  95.04 &  96.50 &
  \blue{77.73} &  \blue{92.97} &  84.74 &  \blue{88.67} &  70.03 &
  86.70 &  81.41 &  83.97 \\
  
  &  ST-Trans \cite{tong2024strans} &  TGRS 2024 &  92.73 &  97.75 &  95.52 &  96.62 &
  76.02 &  89.96 &  85.18 &  87.50 &  70.04 &
  \blue{88.21} &  80.01 &  83.91 \\
 &  SSTNet \cite{chen2024sstnet} &  IEEE TGRS 2024 &  \blue{95.59} &  \blue{98.08} &  98.10 &  \blue{98.09} &
  76.96 &  91.05 &  85.29 &  88.07 &  \blue{71.55} &
  \red{88.56} &  81.92 &  85.11 \\
 &  \textbf{Tridos} (Ours) &  - &
  \red{97.80} &  \red{99.20} &  \blue{99.67} &  \red{99.43} &  \red{80.41} &
  90.71 &  \red{90.60} &  \red{90.65} &  \red{73.72} &  84.49 &  \red{89.35} &
  \red{86.85} \\ \hline
    \end{tabular}}
\end{table*}
\subsubsection{Quantitative Comparison} 
The quantitative results of different detection methods on three datasets are shown in Table \ref{tab:quantitave}.
From this table, we could have {three} obvious findings. 

{Firstly}, model-driven methods usually perform {bad, almost inefficient}. {For example, as the SOTA one, PSTNN can only achieve a low mAP$_{50}$ of 17.31\%, a low Precision of 25.56\% and a low Recall of 68.86\% on DAUB, far lower than any compared data-driven method in TABLE \ref{tab:quantitave}}. One possible reason is that they rely on hand-crafted features and 
{have almost no} adaptively-learning ability to target features.

{The second} is that our Tridos achieves the best performance on most evaluation metrics over three datasets, especially on $\text{mAP}_{50}$ and F1 score. For example, on DAUB dataset, Tridos could obtain the highest $\text{mAP}_{50}$ 97.80\% and the highest F1 99.43\%. In terms of Recall, the 99.67\% by Tridos is only slightly lower than the SOTA 99.79\% by SSTNet. Furthermore, on ITSDT-15K, our Tridos still 
{obtains} the highest $\text{mAP}_{50}$ 80.41\% and F1 90.65\%, far superior to {previous} SOTA $\text{mAP}_{50}$ 76.96\%, and F1 score 88.07\% by SSTNet. Additionally, {it} also {outperforms} the comparison methods almost on all metrics on IRDST.
\begin{figure*}[h]
    \centering
\includegraphics[width=\linewidth]{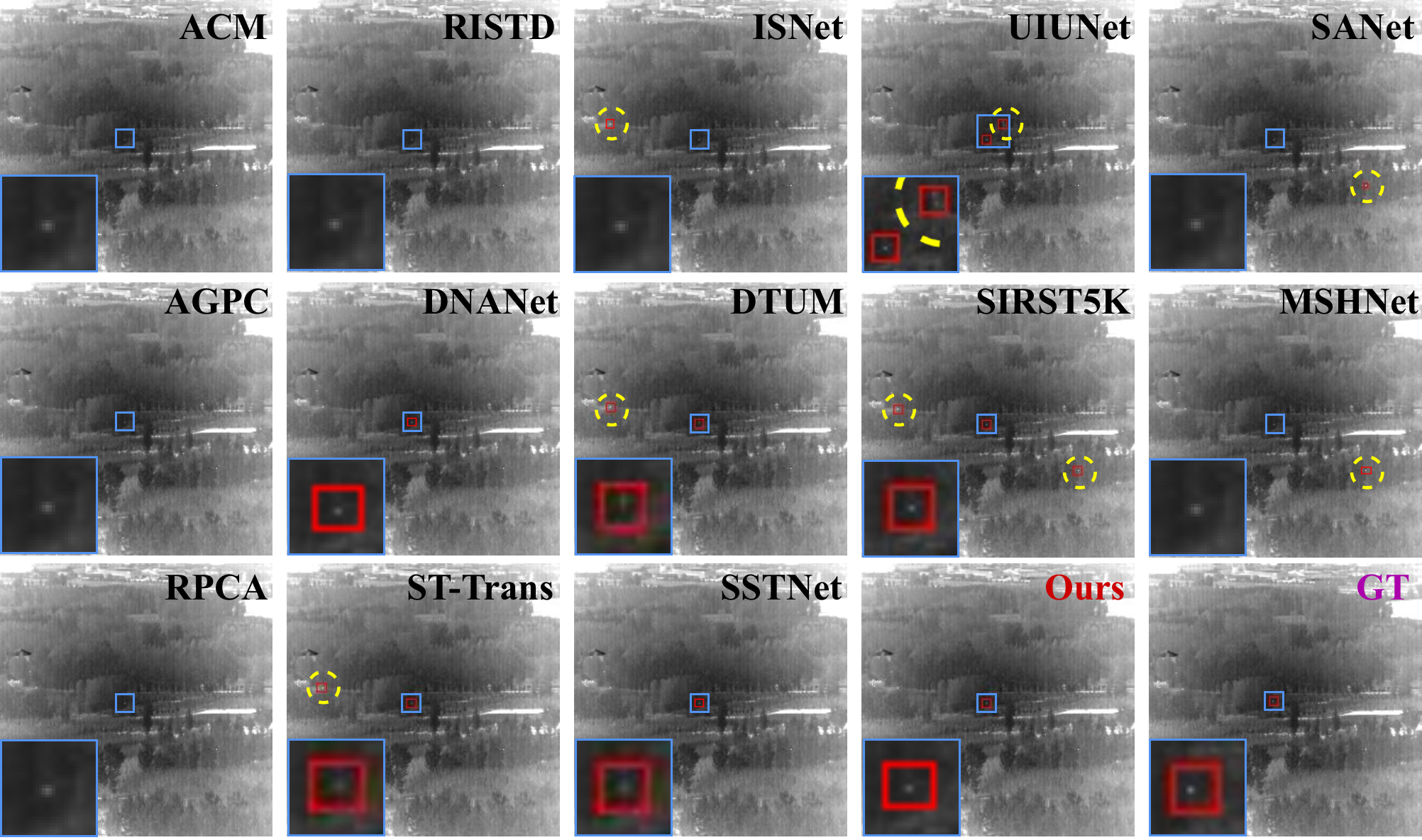}
    \caption{{The visualization comparisons of 14 methods on DAUB, with 21/63.bmp. GT is ground truth. Red and blue boxes represent detected targets and amplified detection regions, respectively. Yellow circles denote false alarms.}}
    \label{fig:vis1}
\end{figure*}
\begin{figure*}[h]
    \centering
\includegraphics[width=\linewidth]{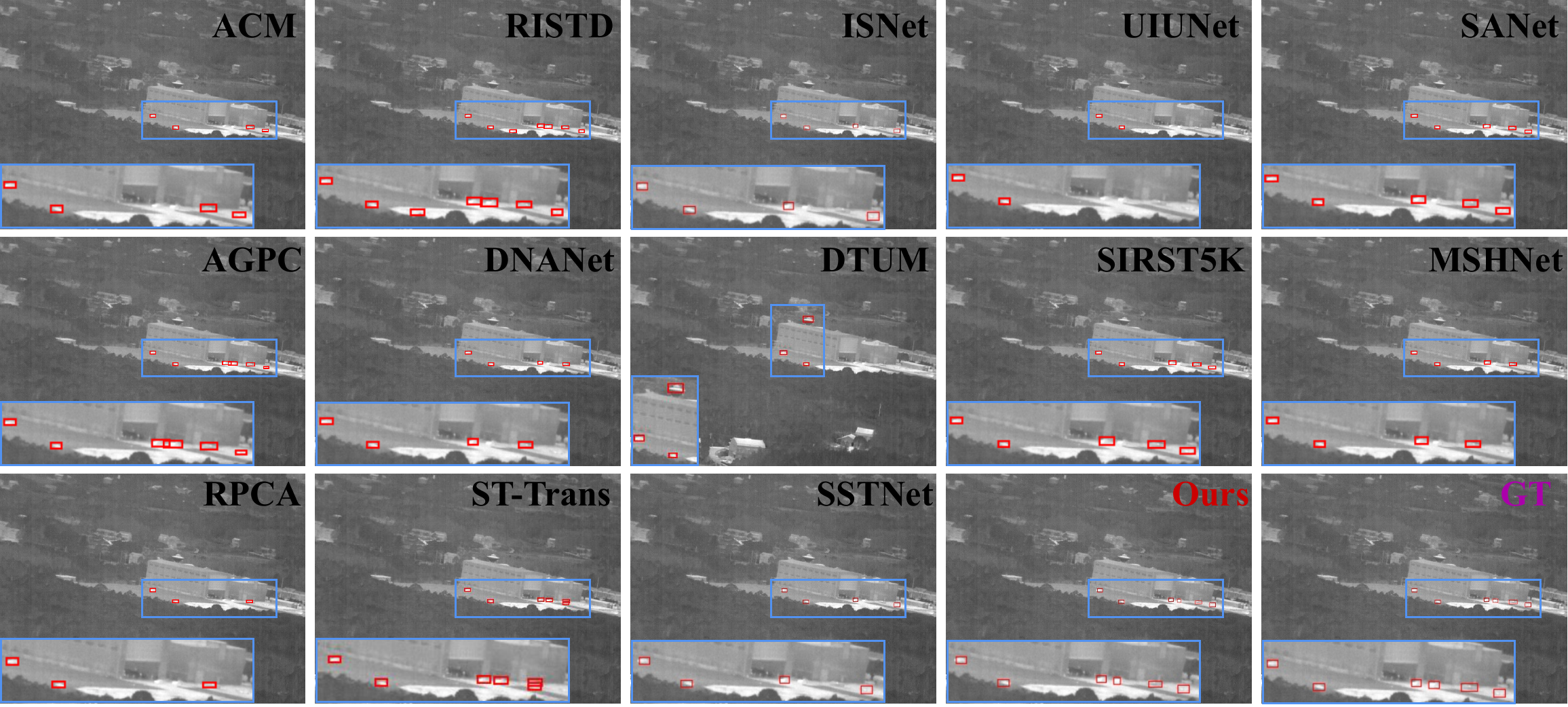}
    \caption{The visualization comparisons of 14 methods on ITSDT-15K, {with 82/205.bmp}. 
{GT is ground truth. Red and blue boxes represent detected targets and amplified detection regions, respectively.}}
    \label{fig:vis2}
\end{figure*}

{The third} is that our Tridos has {stronger} robustness than other ones. Almost all methods {could} perform better on DAUB than on ITSDT-15K and IRDST. For example, {SOTA} SSTNet achieves $\text{mAP}_{50}$ 95.59\% and F1 98.09\% on DAUB, while on ITSDT-15K {its} $\text{mAP}_{50}$ {is} 76.96\% {and} F1 {is} 88.07\%. {Moreover, its} $\text{mAP}_{50}$ is 71.55\% {and} F1 {is} 85.11\% on IRDST. That could be because ITSDT-15K and IRDST contain more complex scenes with the effects of noise and occlusion. 
Nevertheless, on DAUB {our} Tridos {achieves an} $\text{mAP}_{50}$ improvement of 2.21\% and F1 {rise} of 1.34\% {on SSTNet}. On ITSDT-15K and IRDST, the $\text{mAP}_{50}$ increments by Tridos on SSTNet are 3.45\% and 2.17\%, respectively. Meanwhile, on these two datasets, the F1 increments by Tridos on SSTNet are 2.58\% and 1.74\%, respectively. {These comparisons} show that our method could be more tolerant to complex scenarios than others.

{Besides, the comparisons on IRDST indicate that our method achieves higher F1 score than SOTA SSTNet. However, its precision 84.49\% is 4.13\% lower than the 88.56\% by SSTNet under same IOU threshold 0.65. One main reason is that our triple-domain feature learning method tends to detect all targets as many as possible, avoiding miss detection. In real application scenes, low miss detection is usually more important than high false detection. Therefore, our method aims to keep high Recall, by balancing Precision. In fact, if threshold is increased, our Precision could be higher than that of SSTNet. For example, experiments show when threshold is set to 0.7, our Precision is 88.71\%, 0.15\% higher than the 88.56\% achieved by SSTNet.  
}

%comprehensive feature learning strategy, which could include more challenging targets that are not easily distinguishable from background. While increasing recall by identifying more true positives, this strategy increases the number of false positives to some extent, resulting in lower precision.

\subsubsection{Visual Comparison} 
\begin{figure*}[h]
    \centering
\includegraphics[width=\linewidth]{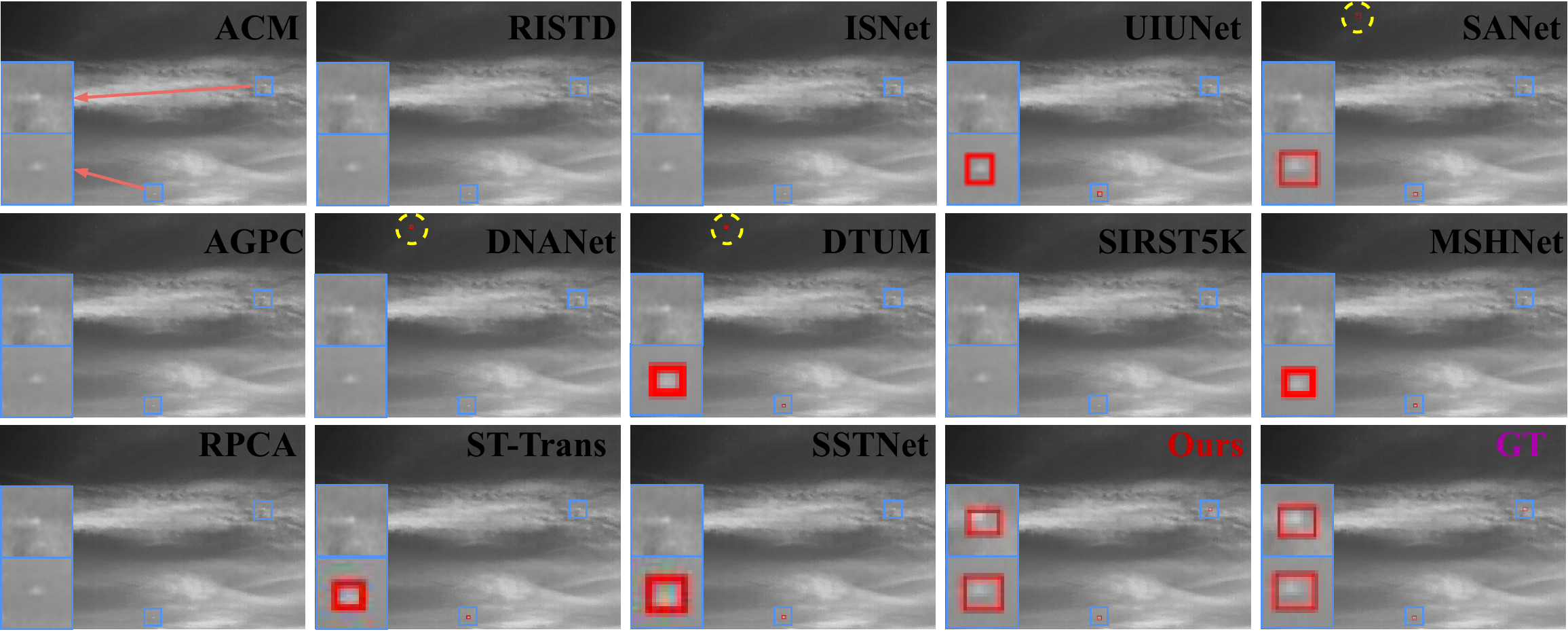}
    \caption{The visualization comparisons of 14 methods on IRDST, {with 6/51.bmp}. {GT is ground truth. Red and blue boxes represent detected targets and amplified detection regions, respectively. Yellow circles denote false alarms.}}
    \label{fig:vis3}
\end{figure*}

We present the visual comparisons of different {detection} methods in Fig. \ref{fig:vis1} - Fig. \ref{fig:vis3}.
%visual-12l
It is evident that our proposed method could usually accurately detect moving infrared small targets. In contrast, other ones often lead to miss detection or false detection. 

For example, in Fig. \ref{fig:vis1}, on DAUB, our Tridos precisely detects the target occluded by {trees}. However, ACM, RISTD, {AGPC and RPCA} cause miss detection. 
{Some methods} incorrectly treat a bright spot as a target, producing false detection, {such as} ISNet, DTUM, SIRST5K and {ST-Trans}.
Moreover, in Fig. \ref{fig:vis2}, on ITSDT-15K, {some methods} fail to detect all targets, such as ACM, DNANet, RPCA and SSTNet. {RISTD} even detects {seven} targets, causing false detection. 
Meanwhile, in Fig. \ref{fig:vis3}, 
on IRDST, SANet, DNANet and DTUM appear false detection. Furthermore, ST-Trans seems to have a smaller bounding box than ground truth.
Some methods {cannot even detect any} targets, such as AGPC, SIRST5K and RPCA. 
% on IRDST, ACM, RISTD, AGPC and DNANet occur miss detection. ISNet appears false detection and it seems to have a bigger bounding box than ground truth.
{In contrast}, our Tridos {could obtain the} bounding boxes with the largest similarity to ground truth.
In summary, we can {observe} that the qualitative results of these visualizations are {much consistent} to the quantitative results in Table \ref{tab:quantitave}, indicating the {superiority} of our method.
%PR curve
\begin{figure*}[h]
    \centering
\includegraphics[width=\linewidth]{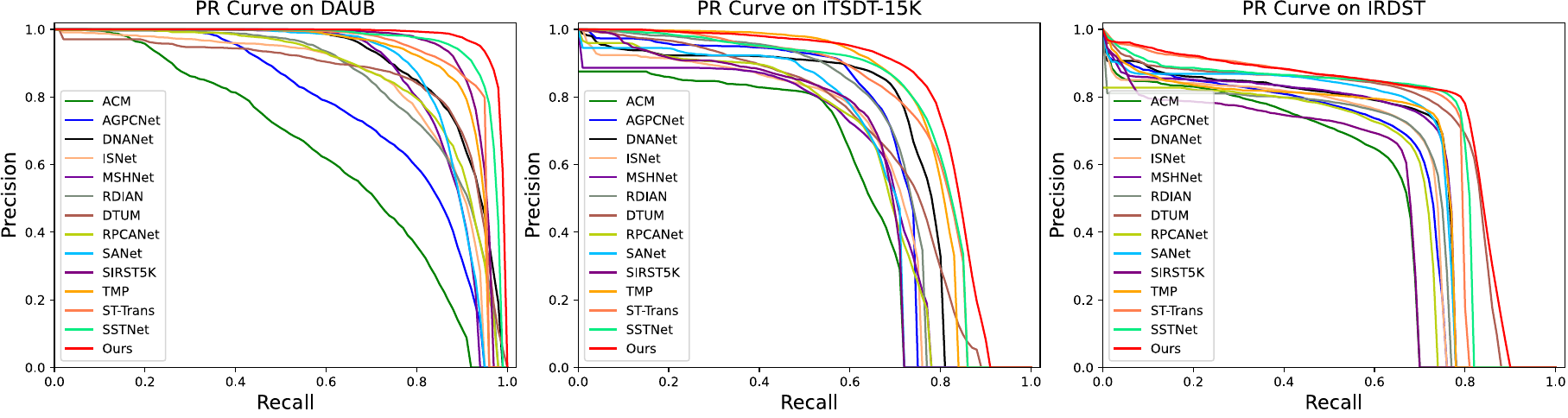}
    \caption{PR curves of {14} representative {detection} methods on datasets DAUB, ITSDT-15K and IRDST.}
    \label{fig:pr}
\end{figure*}
\subsubsection{PR Curve Comparison} 
% PR curve demonstrates the relationship between Precision and Recall and is a comprehensive view of the model's performance.
To evaluate the comprehensive performance of different methods, we draw three groups of Precision-Recall (PR) cures on DAUB, ITSDT-15K and IRDST, as shown in Fig. \ref{fig:pr}. 
By comparison, we could easily observe that our curves {are} almost always above other ones on three datasets. The larger a curve {envelope} area is, the better a method {is}.
Therefore, the three groups of PR curves indicate that our Tridos has the best comprehensive performance with an optimal balance between Precision and Recall.
% Table-Params
\begin{table}[h]
\centering
\caption{
The comparative results {of} model complexity on ITSDT-15K. The best results are marked in bold.
}
\label{tab:params}
\resizebox{\linewidth}{!}{
\begin{tabular}{c|c|cc|ccc}
\hline
\textbf{Methods} & \textbf{Frames} & \textbf{$\text{mAP}_{\textbf{50}}$} $\uparrow$ & \textbf{F1} $\uparrow$ & \textbf{Params} $\downarrow$ & \textbf{GFlops} $\downarrow$ & \textbf{FPS} $\uparrow$ \\ \hline
ACM \cite{NUAAACM}         & 1 & 55.38          & 74.88          & 3.04M          & \textbf{24.73} & \textbf{29.11} \\
RISTD \cite{hou2021ristdnet}        & 1 & 60.47          & 77.93          & 3.28M          & 76.28          & 10.21          \\
ISNet \cite{ISNetzhang2022isnet}       & 1 & 62.29          & 79.18          & 3.49M          & 265.73         & 11.20          \\
UIUNet \cite{wu2022uiuNet}      & 1 & 65.15          & 81.13          & 53.06M         & 456.70         & 3.63           \\
SANet \cite{zhu2023SAnet}       & 1 & 62.17          & 78.64          & 12.40M         & 42.04          & 10.55          \\
AGPCNet \cite{AGPCNet}     & 1 & 67.27          & 82.16          & 14.88M         & 366.15         & 4.79           \\
RDIAN \cite{RDIAN}       & 1 & 68.49          & 82.68          & \textbf{2.74M} & 50.44          & 20.52          \\
DNANet \cite{DNAnetli2022dense}      & 1 & 70.46          & 84.46          & 7.22M          & 135.24         & 4.82           \\
SIRST5K \cite{lu2024sirst5k}     & 1 & 61.52          & 78.36          & 11.48M         & 182.61         & 7.37           \\
MSHNet \cite{liu2024MSH}      & 1 & 60.82          & 77.64          & 6.59M          & 69.59          & 18.55          \\
RPCANet \cite{wu2024rpcanet}      & 1 & 62.28          & 79.22          & 3.21M          & 382.69         & 15.89          \\
DTUM  \cite{li2023directiondtum} & 5 & 67.97          & 82.79          & 9.64M          & 128.16         & 14.28          \\
 TMP  \cite{zhu2024tmp} & 5 & 77.73          & 88.67          & 16.41M          & 92.85         & 6.91          \\
 ST-Trans \cite{tong2024strans} & 5 & 76.02          & 87.50         & 38.13M          & 145.16         & 3.90          \\
SSTNet \cite{chen2024sstnet}      & 5 & 76.96          & 88.07          & 11.95M         & 123.60         & 7.37           \\
Ours         & 5 & \textbf{80.41} & \textbf{90.65} & 14.13M         & 130.72         & 13.71        \\ \hline
\end{tabular} 
}
\end{table}
\subsubsection{Model Complexity Comparison} 
We compare the model complexity of our Tridos with {fifteen} representative methods in Table \ref{tab:params}, mainly {including} \emph{Params}, \emph{GFlops} and \emph{FPS}. In {this} table, we could obviously have two discoveries.

One is that our Tridos has a slight increase in \emph{Params} and \emph{GFlops}. For example, the smallest \emph{Params} in multi-frame methods is 9.64M by DTUM. the best \emph{GFlops} in multi-frame methods is {92.85} by {TMP}.
Our Tridos has a medium \emph{Params} 14.13M and \emph{GFlops} 130.72.
One probable reason is that {DTUM and TMP} are based on spatio-temporal domains, {while} Tridos introduces frequency domain, besides spatio-temporal domains. {
It could be because that converting spatio-domain data into frequency domain involves some additional computations, such as the Fourier transform.
Moreover, {our} network needs extra convolutional layers to coordinate frequency domain features and integrating different domain features further raises \emph{Params} and \emph{GFlops}.
Nevertheless, achieving} obvious performance gains, these complexity costs are acceptable and worthwhile. 

The other is that computational cost increase results in a middle inference speed {of} FPS 13.71. Besides, each multi-frame method {will process} five frame images {in} inference, {so} its FPS is {often} lower than {that of a} single-frame one, e.g., ACM, RDIAN, MSHNet and RPCANet.
Nevertheless, our FPS still seems higher than {that by} RISTD, ISNet, UIUNet, AGPCNet, DNANet, SIRST5K, {ST-Trans} and SSTNet.

% Table-component
\begin{table*}[h]
\centering
\caption{Ablation study on {4} different components, including MSRM, TDEM, LGFM and RCU, on two datasets. RCU(A) only uses $\text{RCU}_1$, {while} RCU(B) uses {both} $\text{RCU}_1$ and $\text{RCU}_2$. RCU(C) uses all RCU {components}.}
\label{tab:ablation1}
\resizebox{\linewidth}{!}{
\begin{tabular}{l|cccccc|cccc|cccc}
\hline
\multirow{2}{*}{\textbf{Settings}} &
  \multirow{2}{*}{\textbf{MSRM}} &
  \multirow{2}{*}{\textbf{TDEM}} &
  \multirow{2}{*}{\textbf{LGFM}} &
  \multirow{2}{*}{\textbf{RCU (A)}} &
  \multirow{2}{*}{\textbf{RCU (B)}} &
  \multirow{2}{*}{\textbf{RCU (C)}} &
  \multicolumn{4}{c}{\textbf{DAUB}} &
  \multicolumn{4}{c}{\textbf{ITSDT-15K}} \\ \cline{8-15}
 &   &   &   &   &   &   &
  \textbf{$\text{mAP}_{\textbf{50}}$} &  \textbf{Precision} &  \textbf{Recall} &  \textbf{F1} &  \textbf{$\text{mAP}_{\textbf{50}}$} &
  \textbf{Precision} &  \textbf{Recall} &  \textbf{F1} \\ \hline
w/o All &  - &  - &  - &  - &  - &  - &
  84.43 &  89.73 &  95.45 & 92.50 &  71.95 &  83.43 &  87.35 &  85.34 \\
w MSRM &  \checkmark &  - &  - &  - &  - &  - &
  92.17 &  94.92 &  93.86 &  94.39 &  75.36 &  83.57 &  91.19 &  87.21 \\
w MSRM \& TDEM &  \checkmark &  \checkmark &  - &  - &  - &  - &
  93.73 &  94.51 &  96.24 &  95.37 &  76.77 &  87.03 &  89.72 &  88.35 \\
w/o RCU &  \checkmark &  \checkmark &  \checkmark &  - &  - &  - &
  95.61 &  96.81 &  97.37 &  97.09 &  78.24 &  85.83 &
  \textbf{92.06} &  88.84 \\
w RCU(A) &  \checkmark &  \checkmark &  \checkmark &  \checkmark &  - &  - &
  95.74 &  97.61 &  97.12 &  97.36 &  79.08 &
  89.66 &  89.50 &  89.58 \\
w RCU(B) &  \checkmark &  \checkmark &
  \checkmark &  - &  \checkmark &  - &  96.25 &  98.00 &  99.33 &  98.66 &  79.68 &
  89.36 &  90.10 &  89.73 \\
w RCU(C) &  \checkmark &  \checkmark &  \checkmark &  - &  - &  \checkmark &
  \textbf{97.80} &  \textbf{99.20} &  \textbf{99.67} &  \textbf{99.43} &
  \textbf{80.41} &  \textbf{90.71} &  90.60 &  \textbf{90.65} \\ \hline
\end{tabular}
}
\end{table*}
\subsection{Ablation Study}
\subsubsection{Effects of Different Components}
We conduct four groups of ablation studies on DAUB and ITSDT-15K to investigate the {effects} of each component {on} our Tridos, as shown in Tables \ref{tab:ablation1} - \ref{tab:fstb1}.

{In} Table \ref{tab:ablation1}, we could have two findings.
{One} is that each component improves {detection} performance to some extent. For example, on DAUB the baseline without any components {only} acquires $\text{mAP}_{50}$ 84.43\% and F1 92.50\%. 
{Once} MSRM {is applied}, $\text{mAP}_{50}$ increases from 84.43\% to 92.17\%, and F1 rises from 92.50\% to 94.39\%. {with} TDEM, $\text{mAP}_{50}$ and F1 could be further improved to 93.73\% and 95.37\%, respectively.
Furthermore, adding frequency domain (w/o RCU) could also provide {an obvious} gain, with $\text{mAP}_{50}$ and F1 respectively rising to 95.61\% and 97.09\%. Besides, Tridos performs variably with different kinds of RCUs (A, B and C).
{The other} is that the best performance could be obtained by applying {all} components together.
%, proving the complementary of each component. 
For example, {with} all components, on DAUB $\text{mAP}_{50}$ and F1 are increased to 97.80\% and 99.43\%, respectively. On ITSDT-15K, $\text{mAP}_{50}$ and F1 are {also} refreshed to 80.41\% and 90.65\%, respectively.

\subsubsection{Effects of NAB and MEU in MSRM}
To explore the potential contribution of NAB and MEU {to} MSRM more deeply, we conduct a group of ablation studies, as shown in Table \ref{tab:msrm}. 
In it, MEU plays an essential role {on} modeling inter-frame spatial relationships. 
For example, with MEU, on DAUB $\text{mAP}_{50}$ and F1 could be increased from 94.92\% to 97.80\%, and 98.19\% to 99.43\%, respectively. It implies the effectiveness of {adopted} memory mechanism.
Similarly, on ITSDT-15K, NAB could improve $\text{mAP}_{50}$ from 78.49\% to 80.41\%, and F1 {is raised} from 89.08\% to 90.65\%.
{These effects} could be attributed to the ability of NAB to capture long-distance dependencies.
%Table-MSRM
\begin{table}[h]
\centering
\caption{{Ablation study on NAB and MEU of MSRM.}}
\label{tab:msrm}
\resizebox{\linewidth}{!}{
\begin{tabular}{l|cccc|cccc}
\hline
\multirow{2}{*}{\textbf{Settings}} &
  \multicolumn{4}{c}{\textbf{DAUB}} &
  \multicolumn{4}{c}{\textbf{ITSDT-15K}} \\ \cline{2-9}
 &
  \textbf{$\text{mAP}_{\textbf{50}}$} &
  \textbf{Precision} &
  \textbf{Recall} &
  \textbf{F1} &
  \textbf{$\text{mAP}_{\textbf{50}}$} &
  \textbf{Precision} &
  \textbf{Recall} &
  \textbf{F1} \\ \hline
MSRM w/o NAB &  96.53 &  98.24 &  99.52 &  98.88 &  78.49 &  88.03 &  90.16 &  89.08 \\
 MSRM w/o MEU &  94.92 &  97.57 &  98.81 &  98.19 &  74.05 &  84.91 &  88.74 &  86.78 \\
MSRM w/o All &  93.47 &  95.33 &  99.04 &  97.15 &  73.16 &  85.67 &  86.29 &  85.97 \\
MSRM  &  \textbf{97.80} &  \textbf{99.20} &  \textbf{99.67} &
  \textbf{99.43} &  \textbf{80.41} &  \textbf{90.71} &  \textbf{90.60} &  \textbf{90.65} \\ \hline
\end{tabular}}
\end{table}

{To verify the motivation {of} devising NAB and distinguish it from self-attention mechanism, we perform a group of ablation experiments, as shown in Table \ref{tab:self}. `Conv' represents {a} baseline {adopting} convolutional layers {rather than} attention blocks.
We observe that attention mechanism {could} improve baseline performance. 
Specifically, on DAUB, self-attention {could} improve $\text{mAP}_{50}$ from 94.98\% to 96.64\% and F1 from 97.12\% to 98.59\%.
NAB could further raise $\text{mAP}_{50}$ to 97.80\% and F1 to 99.43\%.
{In detail}, self-attention catches the dependencies in sequences by computing the attention scores of all position pairs. In contrast, NAB focuses more on capturing comprehensive contexts and long-distance dependencies.
}
%nab vs selfattention
\begin{table}[h]
\centering
\caption{Ablation study on the effectiveness of NAB and self-attention {module}.}
\label{tab:self}
\resizebox{\linewidth}{!}{
\begin{tabular}{
l|
c 
c 
c 
c|
c 
c 
c 
c } \hline
 & \multicolumn{4}{c}{\textbf{DAUB}}         & \multicolumn{4}{c}{\textbf{ITSDT-15K}}        \\ \cline{2-9} 
\multirow{-2}{*}{\textbf{Settings}} &
  \textbf{$\text{mAP}_{\textbf{50}}$} &
  \textbf{Precision} &
  \textbf{Recall} &
  \textbf{F1} &
  \textbf{$\text{mAP}_{\textbf{50}}$} &
  \textbf{Precision} &
  \textbf{Recall} &
  \textbf{F1} \\ \hline
Conv           & 94.98          & 96.76          & 97.48          & 97.12          & 77.18          & 86.43          & 88.71          & 87.56          \\
Self-attention  & 96.64          & 98.07          & 99.12          & 98.59          & 78.53          & 87.64          & 89.23          & 88.43          \\
NAB             & \textbf{97.80} & \textbf{99.20} & \textbf{99.67} & \textbf{99.43} & \textbf{80.41} & \textbf{90.71} & \textbf{90.60} & \textbf{90.65} \\ \hline
\end{tabular}}
\end{table}

\subsubsection{Effects of ResB in TDEM}
From Table \ref{tab:tdem1}, it is obvious that two residual blocks will {make Triods to} get the best performance. For example, on ITSDT-15K, {the} TDEM without residual blocks (TDEM w/o All) only obtains $\text{mAP}_{50}$ 74.05\% and F1 86.78\%. With ResB1 (TDEM w ResB1), $\text{mAP}_{50}$ and F1 will raise to 76.25\% and 87.69\%, {respectively}. Moreover, with both (TDEM), $\text{mAP}_{50}$ and F1 further increase 4.13\% and 2.27\%, {respectively}. {These improvements} could be because {both} enhance differential information from {multiple} perspectives, {capturing} more representative motion features.
% Table-TDEM
\begin{table}[h]
\centering
\caption{Ablation study on two residual blocks of TDEM.}
\label{tab:tdem1}
\resizebox{\linewidth}{!}{
\begin{tabular}{l|cccc|cccc}
\hline
\multirow{2}{*}{\textbf{Settings}} &
  \multicolumn{4}{c}{\textbf{DAUB}} &
  \multicolumn{4}{c}{\textbf{ITSDT-15K}} \\ \cline{2-9}
 &
  \textbf{$\text{mAP}_{\textbf{50}}$} &
  \textbf{Precision} &
  \textbf{Recall} &
  \textbf{F1} &
  \textbf{$\text{mAP}_{\textbf{50}}$} &
  \textbf{Precision} &
  \textbf{Recall} &
  \textbf{F1} \\ \hline
  TDEM w/o ResB1 &  95.49 &  98.08 &  98.87 &  98.48 &
  77.21 &  86.58 &  90.25 &  88.38 \\
  TDEM w/o ResB2 &  94.45 &  96.24 &  99.42 &  97.80 &
  76.25 &  84.57 &  91.05 &  87.69 \\
TDEM w/o All &  93.89 &  96.37 &  98.87 &
  97.60 &  74.05 &  84.91 &  88.74 &  86.78 \\
TDEM  &  \textbf{97.80} &  \textbf{99.20} &  \textbf{99.67} &  \textbf{99.43} &
  \textbf{80.41} &  \textbf{90.71} &  \textbf{90.60} &  \textbf{90.65} \\ \hline
\end{tabular}}
\end{table}
\subsubsection{{Effects of FT, ConvB, and SwinB in LGFM}}
Additionally, we investigate the effects of \emph{Fourier Transform} (FT), \emph{Conv Branch} (ConvB) and \emph{Swin Transformer Branch} (SwinB) in Table \ref{tab:fstb1}.
From it, we could have {two} apparent findings. {First}, the global contexts acquired by SwinB is as valuable as the local features obtained by ConvB. For example, on DAUB, without ConvB (LGFM w/o ConvB), {it} obtains {an $\text{mAP}_{50}$} 95.95\% and {an F1} 98.72\%. Without SwinB, {it} achieves {an $\text{mAP}_{50}$} 95.76\% and {an F1} 98.39\%.
{Second}, frequency transformation is a crucial strategy for detecting small targets.
For example, on ITSDT-15K, ``LGFM w/o FT" only acquires {an $\text{mAP}_{50}$} 73.54\% and {an F1} 85.47\%, even {assembled} with 
{both} branches. It could prove that our LGFM is a successful strategy for perceiving frequency information.
% Table-LGFM
\begin{table}[h]
\centering
\caption{Ablation study of Fourier transform (FT), the ConvB and SwinB of LGFM.}
\label{tab:fstb1}
\resizebox{\linewidth}{!}{
\begin{tabular}{l|cccc|cccc}
\hline
\multirow{2}{*}{\textbf{Settings}} &
  \multicolumn{4}{c}{\textbf{DAUB}} &
  \multicolumn{4}{c}{\textbf{ITSDT-15K}} \\ \cline{2-9}
 &  \textbf{$\text{mAP}_{\textbf{50}}$} &  \textbf{Precision} &  \textbf{Recall} &  \textbf{F1} &
  \textbf{$\text{mAP}_{\textbf{50}}$} &  \textbf{Precision} &  \textbf{Recall} &  \textbf{F1} \\ \hline
LGFM w/o ConvB &  95.95 &  98.61 &  98.94 &  98.72 &  76.73 &  86.04 &  90.35 &  88.14 \\
LGFM w/o SwinB &  95.76 &  97.52 &  99.27 &  98.39 &  74.77 &  82.01 &  92.37 &  86.88 \\
LGFM w/o FT &  94.27 &  97.46 &  97.60 &  97.53 &  73.54 &  86.38 &  84.59 &  85.47 \\
LGFM w/o All &  92.93 &  97.14 &  96.45 &  96.80 &  72.90 &  80.64 &  89.61 &  84.89 \\
LGFM w All &  \textbf{97.80} &  \textbf{99.20} &  \textbf{99.67} &  \textbf{99.43} &
  \textbf{80.41} &  \textbf{90.71} &  \textbf{90.60} &  \textbf{90.65} \\ \hline
\end{tabular}}
\end{table}

\begin{figure}[h]
    \centering
\includegraphics[width=\linewidth]{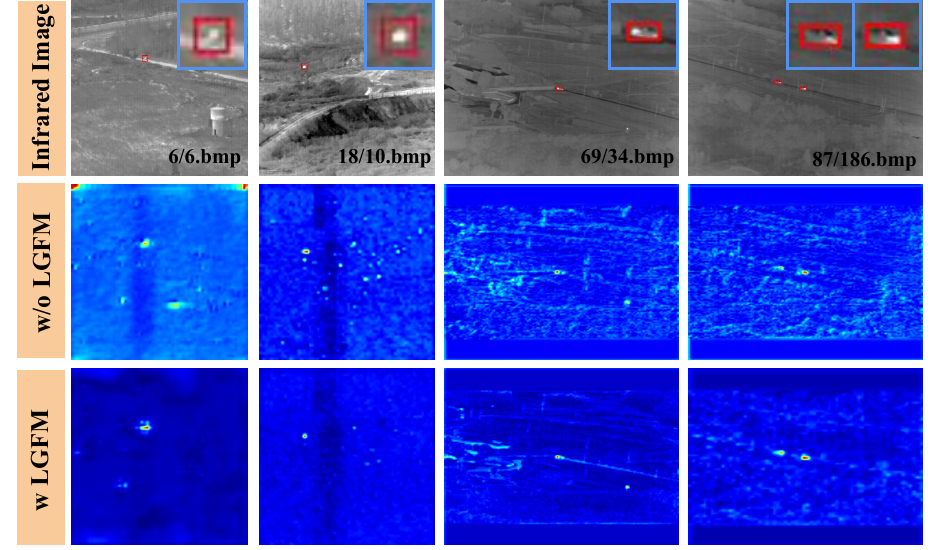}
    \caption{{Feature visualization comparisons before (w/o) and after (w) LGFM. The first two columns are on DAUB, and the rest are on ITSDT-5K.}}
    \label{fig:heat}
\end{figure}

{To demonstrate the effectiveness of frequency learning, we select four samples from DAUB and ITSDT-15K, and visualize the features before and after LGFM, as shown in Fig. \ref{fig:heat}. In it, we could have two findings. One is that the feature map by ``w LGFM'' eliminates some background noise interference. This demonstrates that LGFM could effectively filter irrelevant information.
{The other} is that after applying LGFM, the feature response to targets is obviously stronger. This comparison implies that LGFM could make our network more focused on target regions.
}
% accept 15K

\subsubsection{{Effects of CAB and SAB in RCU}}
{We conduct a group of experiments to validate the effectiveness of CAB and SAB.
In Table \ref{tab:cab}, we could find that integrating CAB and SAB can promote RCU performance, acquiring the best results.
For example, on ITSDT-15K, SAB could increase $\text{mAP}_{50}$ from 76.73\% to 77.75\%, and F1 from 88.25\% to 88.88\%. CAB could boost $\text{mAP}_{50}$ to 78.61\% and F1 to 89.03\%. Detection performance will peak, i.e., $\text{mAP}_{50}$ 80.41\% and F1 90.65\%, only when both CAB and SAB are employed.
This is because when CAB and SAB are combined, they could complement each other by capturing spatial and channel dependencies to achieve more robust feature representations.
}
% CAB SAB
\begin{table}[h]
\centering
\caption{Ablation study on the CAB and SAB of RCU.}
\label{tab:cab}
\resizebox{\linewidth}{!}{
\begin{tabular}{
l |
c 
c 
c 
c|
c 
c 
c 
c } \hline
 & \multicolumn{4}{c}{\textbf{DAUB}}         & \multicolumn{4}{c}{\textbf{ITSDT-15K}}        \\\cline{2-9}
\multirow{-2}{*}{\textbf{Settings}} &
  \textbf{$\text{mAP}_{\textbf{50}}$} &
  \textbf{Precision} &
  \textbf{Recall} &
  \textbf{F1} &
  \textbf{$\text{mAP}_{\textbf{50}}$} &
  \textbf{Precision} &
  \textbf{Recall} &
  \textbf{F1} \\ \hline
  RCU w/o all              & 92.41          & 94.24          & 99.12          & 96.62          & 76.73          & 87.00          & 89.54          & 88.25          \\
RCU w SAB              & 94.20          & 96.67          & 99.37          & 97.69          & 77.55          & 86.94          & 90.90          & 88.88          \\
RCU w CAB              & 95.64          & 97.35          & 99.58          & 98.45          & 78.61          & 88.23          & 89.85          & 89.03          \\

RCU w All                     & \textbf{97.80} & \textbf{99.20} & \textbf{99.67} & \textbf{99.43} & \textbf{80.41} & \textbf{90.71} & \textbf{90.60} & \textbf{90.65}\\ \hline
\end{tabular}}
\end{table}

\subsubsection{Effects of Time Window Size $T$}
\begin{figure}[h]
    \centering
\includegraphics[width=\linewidth]{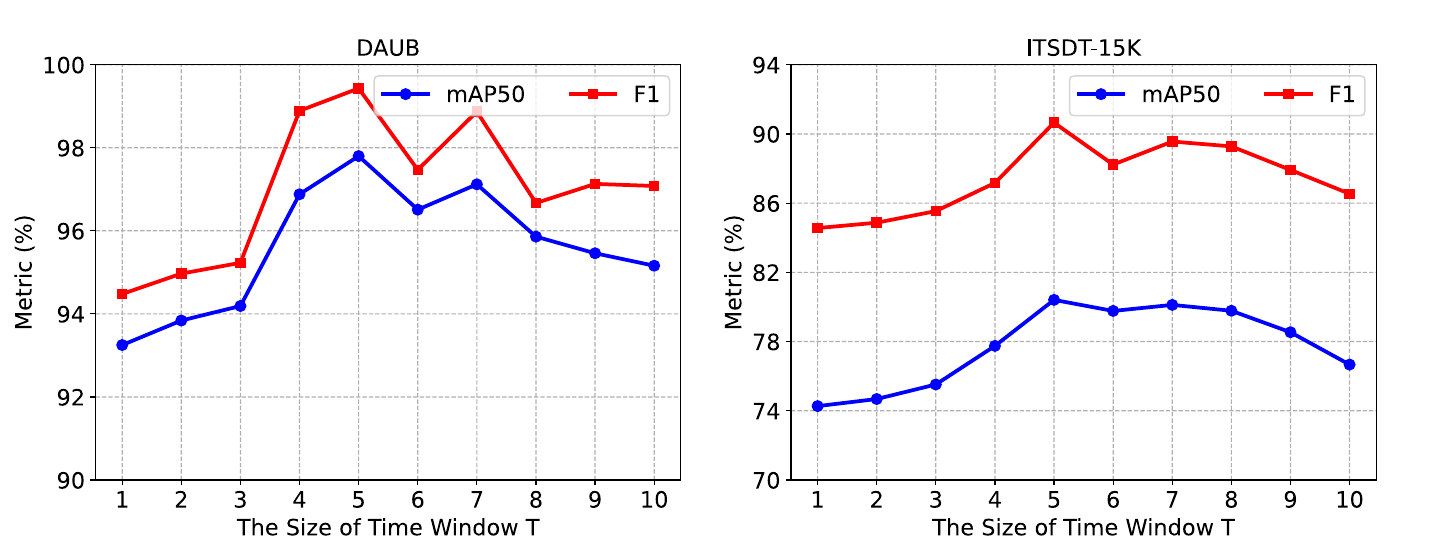}
    \caption{The effects of time window {size} $T$ on Tridos.}
    \label{fig:images}
\end{figure}

We conduct a group of experiments with different time window sizes to investigate the impact of time window $T$ on detection performance, as shown in Fig. \ref{fig:images}. 

Time window {size} $T$ could provide contexts for {a} keyframe.
In {Fig. \ref{fig:images}}, we could {observe} {two} findings. {First}, our method obtains the peak detection performance on two datasets when $T=5$. {Second}, proper {$T$} is crucial for infrared small target detection. 
For example, when $T\leq3$, increasing $T$ {will result} in less gain. when $T>3$, detection performance {increment is} more evident. Furthermore, when $T > 5$ a further increase {of} $T$ will lead to fluctuation, even a decrease {on} metrics.
This could be {explained by the} fact that capturing the temporal contexts of consecutive frames would be insufficient if {$T$} is too small. Conversely, if {it} is too {big}, there would be a risk of excessive redundant {features}, causing interference.
{Considering this, we always set} $T = 5$ in experiments.

\subsubsection{Effects of {Different Regression} Losses} 
To validate the impacts of {$\mathcal{L}_{dvr}$}, we further design a group of comparison experiments {with three different regression loss functions}, as shown in Table \ref{tab:loss}. 
In {this} table, ``$\mathcal{L}_{iou}$'' indicates that the loss term $\mathcal{L}_{reg}$ in Eq. (\ref{eqloss}) is {set to} $\mathcal{L}_{iou}$. ``$\mathcal{L}_{nwd}$'' denotes that $\mathcal{L}_{reg}$ is {set to} $\mathcal{L}_{nwd}$. {Moreover}, ``$\mathcal{L}_{dvr}$'' represents that $\mathcal{L}_{reg}$ is {set to} $\mathcal{L}_{dvr}$, {computed by} Eq. (\ref{eqab}).

By comparison, our dual-view regression loss, {i.e.,  $\mathcal{L}_{dvr}$}, could gain {a} peak performance {on each dataset}. 
For example, on DAUB, $\mathcal{L}_{iou}$ only realizes 96.15\% and 98.56\% on $\text{mAP}_{50}$ and F1, {respectively}. 
{In contrast}, $\mathcal{L}_{dvr}$ could get 97.80\% and 99.43\% on $\text{mAP}_{50}$ and F1, {respectively}. 
One possible {explanation} is that our $\mathcal{L}_{dvr}$ could improve detection accuracy by minimizing the distribution differences of bounding boxes.
%, {although} it is sensitive to noise, which could result in inaccurate target localization.
In summary, our dual-view regression loss {computed on $\mathcal{L}_{iou}$ and $\mathcal{L}_{nwd}$} incorporates the advantages of balancing small target localization and bounding box distribution.	
%Table-loss
\begin{table}[h]
\centering
\caption{Comparisons of {three regression} loss functions.}
\label{tab:loss}
\resizebox{\linewidth}{!}{
\begin{tabular}{l|cccc|cccc}
\hline
\multirow{2}{*}{\textbf{Settings}} &
  \multicolumn{4}{c}{\textbf{DAUB}} &
  \multicolumn{4}{c}{\textbf{ITSDT-15K}} \\ \cline{2-9}
 &
  \textbf{$\text{mAP}_{\textbf{50}}$} &
  \textbf{Precision} &
  \textbf{Recall} &
  \textbf{F1} &
  \textbf{$\text{mAP}_{\textbf{50}}$} &
  \textbf{Precision} &
  \textbf{Recall} &
  \textbf{F1} \\ \hline
$\mathcal{L}_{iou}$ & 96.15 & 98.09 & 99.04 & 98.56 & 78.20 & 86.71 & 89.85 & 88.25 \\
$\mathcal{L}_{nwd}$ & 93.66 & 96.45 & 98.54 & 97.48 & 73.87 & 82.17 & 84.45 & 83.29 \\
$\mathcal{L}_{dvr}$ &
  \textbf{97.80} &
  \textbf{99.20} &
  \textbf{99.67} &
  \textbf{99.43} &
  \textbf{80.41} &
  \textbf{90.71} &
  \textbf{90.60} &
  \textbf{90.65} \\ \hline
\end{tabular} }
\end{table}

\subsubsection{Effects of Different Hyper-parameters}
\begin{figure}[h]
    \centering
\includegraphics[width=\linewidth]{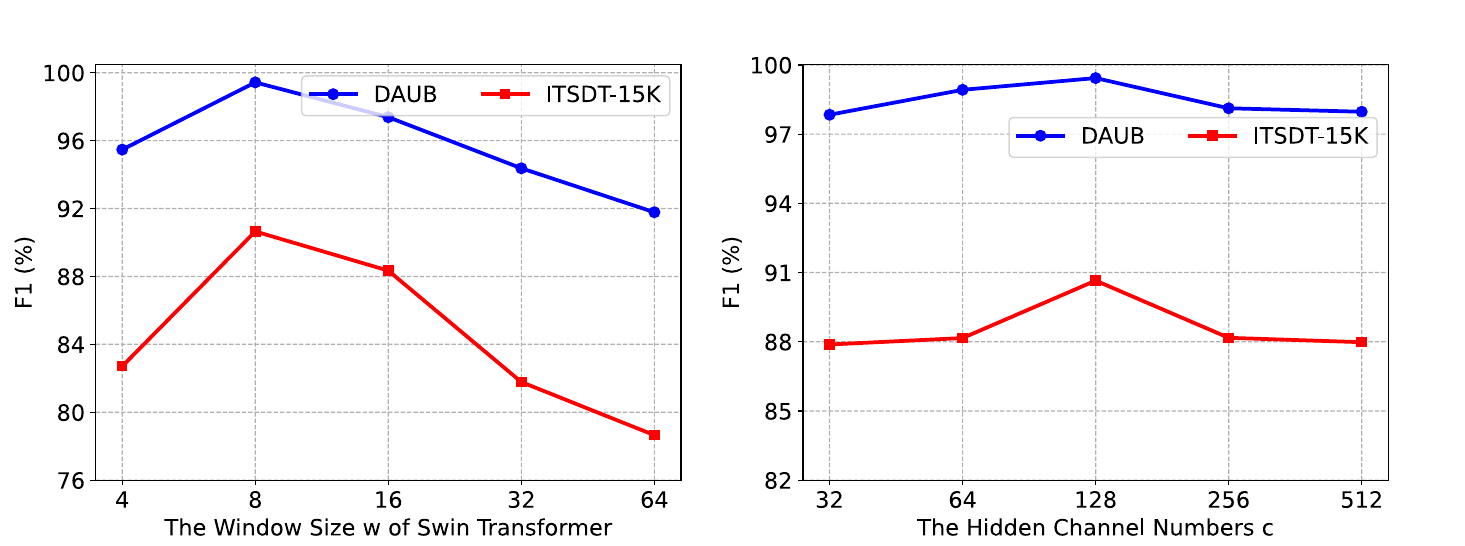}
    \caption{The effects of hyper-parameters $w$ and $c$ on Tridos.}
    \label{fig:hyper}
\end{figure}

As shown in Fig. \ref{fig:hyper}, we design two groups of experiments to evaluate the effects of hyper-parameters $w$ (Swin Transformer window size) and $c$ (hidden channel number) on Tridos. 
It {is seen} that F1 {could reach to} a peak on DAUB and ITSDT-15K when $w=8$. Similar to this, the highest value of F1 {could be} obtained when $c = 128$. 
Besides, we could find that Tridos's detection performance is more sensitive to $w$ than {to} $c$.
Overall, the optimal {$w$ and $c$} are {often} 8 and 128, respectively.

{Furthermore, we conduct a group of experiments to analyze the effects of two {additional} hyper-parameters $\alpha$ and $\beta$ in Eq. (\ref{eqab}).
As shown in Table \ref{tab:beta}, we change $\alpha$ and $\beta$ to different settings,  while keep other parameters fixed.
In this table, it is evident that $\alpha$ and $\beta$ obviously impact the detection performance of our Tridos. When $\alpha = 0.5$ and $\beta = 0.5$, detection performance will peak. According to Eq. (\ref{eqab}), it implies that 
$\mathcal{L}_{iou}$ and $\mathcal{L}_{nwd}$ have the same importance in promoting the detection performance of our Tridos.  
}
\begin{table}[h]
\centering
\caption{The effects of hyper-parameters $\alpha$ and $\beta$.}
\label{tab:beta}
\resizebox{\linewidth}{!}{
\begin{tabular}{l |c c c c |c c c c } \hline
 & \multicolumn{4}{c}{\textbf{DAUB}}         & \multicolumn{4}{c}{\textbf{ITSDT-15K}} \\ \cline{2-9}
\multirow{-2}{*}{\textbf{\begin{tabular}[c]{@{}c@{}}Settings\\ ($\boldsymbol{\alpha}$, $\boldsymbol{\beta}$)\end{tabular}}} &
  \textbf{$\text{mAP}_{\textbf{50}}$} &
  \textbf{Precision} &
  \textbf{Recall} &
  \textbf{F1} &
  \textbf{$\text{mAP}_{\textbf{50}}$} &
  \textbf{Precision} &
  \textbf{Recall} &
  \textbf{F1} \\ \hline
(0.1, 0.9)& 94.88          & 96.74              & 99.03           & 97.88          & 76.94          & 84.40              & 92.20           & 88.13          \\
(0.3, 0.7) & 96.27          & 98.17              & 99.06           & 98.61          & 79.87          & 90.54              & 89.50           & 90.02          \\
(0.5, 0.5) & \textbf{97.80} & \textbf{99.20}     & \textbf{99.67}  & \textbf{99.43} & \textbf{80.41} & \textbf{90.71}              & \textbf{90.60}  & \textbf{90.65} \\
(0.7, 0.3) & 95.61          & 97.42              & 99.23           & 98.32          & 77.63          & 88.64              & 89.15           & 88.89          \\
(0.9, 0.1) & 95.16          & 96.99              & 99.42           & 98.19          & 77.41          & 86.78              & 90.84           & 88.76   \\ \hline

\end{tabular}}
\end{table}

\subsubsection{Comparison with Swin Transformer and YOLOX} 
To further illustrate the effectiveness of our method, we choose conventional Swin-Transformer {(SwinT)} and YOLOX {for performance comparisons}, as shown in Table \ref{tab:swinyolo}.

{It could be found} that our Tridos with frequency-aware and memory enhancement achieves evidently higher detection metrics on {each} dataset. 
For example, on ITSDT-15K, Tridos could acquire an $\text{mAP}_{50}$ 80.41\%, Precision 90.71\%, Recall 90.60\%, and F1 90.65\%, {far superior to} SwinT and YOLOX. 
This group of {experiments} proves that {our superiority mainly attributed to proposed triple-domain feature learning, not only to the simple combination of} SwinT and YOLOX.
%Table-swinyolo
\begin{table}[h]
\centering
\caption{Comparisons with SwinT and YOLOX.}
\label{tab:swinyolo}
\resizebox{\linewidth}{!}{
\begin{tabular}{l|cccc|cccc}
\hline
\multirow{2}{*}{\textbf{Methods}} &
  \multicolumn{4}{c}{\textbf{DAUB}} &
  \multicolumn{4}{c}{\textbf{ITSDT-15K}} \\ \cline{2-9}
 &
  \textbf{$\text{mAP}_{\textbf{50}}$} &
  \textbf{Precision} &
  \textbf{Recall} &
  \textbf{F1} &
  \textbf{$\text{mAP}_{\textbf{50}}$} &
  \textbf{Precision} &
  \textbf{Recall} &
  \textbf{F1} \\ \hline
SwinT & 83.40 & 90.46 & 93.45 & 91.93 & 49.93 & 78.56 & 64.30 & 70.72 \\
YOLOX & 85.62 & 90.73 & 94.75 & 92.70 & 72.15 & 84.43 & 86.85 & 85.62 \\
\textbf{Ours} &
  \textbf{97.80} &
  \textbf{99.20} &
  \textbf{99.67} &
  \textbf{99.43} &
  \textbf{80.41} &
  \textbf{90.71} &
  \textbf{90.60} &
  \textbf{90.65} \\ \hline
\end{tabular}}
\end{table}

\section{Conclusions}
For extending feature learning, this paper proposes {Tridos} to capture the saptio-temporal-frequency features of moving infrared small targets.
% from the view of spatial-temporal and frequency domains.
% even if targets are moving fast or in complex scenarios.
% We propose a strategy of memory enhancement to imitate the human visual system and design a local-global frequency-aware block to reduce the effects of clutter and motion blur. Besides, the residual compensation unit (RCU) could coordinate the features of different domains.
% This could be attributed to the perception of frequency domain information.
It enhances spatio-temporal feature representation through frequency domain and models inter-frame spatial relationships by memory enhancement mechanism.
% , which makes the network robust against noise and clutter .
%Furthermore, the strategy of memory enhancement to imitate the human visual system could accurately model inter-frame spatial relationships. 
% Besides, the features of different domains need to be coordinated to eliminate mismatches.
To comprehensively evaluate detection methods, an additional dataset ITSDT-15K {is collected}. 
The comparison experiments on DAUB, IRDST and {ITSDT-15K} prove the superiority of {our} Tridos to existing SOTA methods. 
% It could often outperform existing SOTA methods, with medium computing cost and inferring speed. 
Moreover, ablation studies further verify the effectiveness and merits of all elaborately-designed components, {including MSRM, TDEM, LGFM and RCU} in our Tridos.
%{Although achieving expected detection performance, using frequency domain will inevitably increase network parameters and computation costs. Additionally, integrating frequency information could introduce complexity in the training process, requiring more advanced optimization techniques to ensure convergence.
%In the future, we would explore more effective and lightweight triple-domain detection schemes to solve these limitations.}
{One obvious weakness} of our method is that it could also be {with high model complexity} and computation cost, {just as most data-driven ones}. In future work, {how to} efficiently learn infrared small target features {by lightweight neural networks} in spatio-temporal-frequency domains {is worthy of further exploring}.

\bibliographystyle{IEEEtran}
%\bibliographystyle{elsarticle-harv}  %这里不用改，对应的elsarticle-num.bst apalike文件
%\biboptions{authoryear}%这是在正文中的引用方式，使用作者名称及年份
\bibliography{fre}  %填写.bib文件的文件名
\vfill
\end{document}